\documentclass[10pt]{article} 
\usepackage[preprint]{tmlr}

\usepackage{amsmath,amsfonts,bm}

\def\eqref#1{equation~\ref{#1}}

\def\1{\bm{1}}

\DeclareMathAlphabet{\mathsfit}{\encodingdefault}{\sfdefault}{m}{sl}
\SetMathAlphabet{\mathsfit}{bold}{\encodingdefault}{\sfdefault}{bx}{n}

\usepackage{hyperref}
\usepackage{url}

\usepackage[utf8]{inputenc} 
\usepackage[T1]{fontenc}    
\usepackage{hyperref}       
\usepackage{url}            
\usepackage{booktabs}       
\usepackage{amsfonts}       
\usepackage{nicefrac}       
\usepackage{microtype}      
\usepackage{xcolor}         

\usepackage{comment}
\usepackage{amssymb}
\usepackage{amsmath}
\usepackage{graphicx}
\usepackage{subcaption}

\usepackage{multirow}
\usepackage{multicol}
\usepackage{fontawesome}

\usepackage[ruled]{algorithm2e}
\SetKwComment{Comment}{/* }{ */}
\RestyleAlgo{ruled}
\usepackage{amsmath,amssymb}

\usepackage{pifont}
\DeclareSymbolFont{extraup}{U}{zavm}{m}{n}
\DeclareMathSymbol{\varheart}{\mathalpha}{extraup}{86}
\DeclareMathSymbol{\vardiamond}{\mathalpha}{extraup}{87}
\newenvironment{allintypewriter}{\ttfamily}{\par}

\title{Multi-Attribute Constraint Satisfaction via Language Model Rewriting}

\author{Ashutosh Baheti$^{\diamondsuit, \clubsuit,*}$, Debanjana Chakraborty$^{\vardiamond}$, Faeze Brahman$^{\clubsuit}$, Ronan Le Bras$^{\clubsuit}$, \\
Ximing Lu$^{\heartsuit, \clubsuit}$, Nouha Dziri$^{\clubsuit}$, Yejin Choi$^{\heartsuit, \clubsuit}$, Mark Riedl$^{\diamondsuit}$, Maarten Sap$^{\spadesuit, \clubsuit}$ \\
     \textnormal{$^{\diamondsuit}$ Georgia Institute of Technology, $^{\heartsuit}$University of Washington, $^{\vardiamond}$The Ohio State University, \\
     $^{\spadesuit}$Carnegie Mellon University, $^{\clubsuit}$Allen Institute for Artificial Intelligence} \\
     $^*$\texttt{abaheti95@gatech.edu}}

\newcommand{\frameworkLong}{Multi-Attribute Constraint Satisfaction}
\newcommand{\framework}{\textsc{MACS}}
\newcommand{\benchmarkLong}{Fine-grained Constraint Satisfaction}
\newcommand{\benchmark}{\textsc{FineCS}}

\begin{document}

\maketitle
\begin{abstract}

Obeying precise constraints on top of multiple external attributes is a common computational problem underlying seemingly different domains, from controlled text generation to protein engineering.
Existing language model (LM) controllability methods for multi-attribute constraint satisfaction often rely on specialized architectures or gradient-based classifiers, limiting their flexibility to work with arbitrary black-box evaluators and pretrained models.
Current general-purpose large language models, while capable, cannot achieve fine-grained multi-attribute control over external attributes. 
Thus, we create {\frameworkLong} (\framework), a generalized method capable of finetuning language models on any sequential domain to satisfy user-specified constraints on multiple external real-value attributes.
Our method trains LMs as editors by sampling diverse multi-attribute edit pairs from an initial set of paraphrased outputs. During inference, LM iteratively improves upon its previous solution to satisfy constraints for all attributes by leveraging our designed constraint satisfaction reward. We additionally experiment with reward-weighted behavior cloning to further improve the constraint satisfaction rate of LMs. To evaluate our approach, we present a new {\benchmarkLong} ({\benchmark}) benchmark, featuring two challenging tasks: (1) Text Style Transfer, where the goal is to simultaneously modify the sentiment and complexity of reviews, and (2) Protein Design, focusing on modulating fluorescence and stability of Green Fluorescent Proteins (GFP). Our empirical results show that {\framework} achieves the highest threshold satisfaction in both {\benchmark} tasks, outperforming strong domain-specific baselines. Our work opens new avenues for generalized and real-value multi-attribute control, with implications for diverse applications spanning natural language processing and bioinformatics.

\end{abstract}

\section{Introduction}

Multi-attribute constraint satisfaction is a challenging problem that holds many useful applications in the domains of natural language processing (NLP), drug design, and protein engineering. 
In NLP, numerous classifiers and regressors exist for detecting individual linguistic attributes such as fluency, sentiment, formality, and complexity. Enabling {\em fine-grained granular control} over such attributes will allow users to personalize any text with their desired style \citep{NEURIPS2021_79ec2a42, kumar-etal-2022-gradient}.
In the realm of medicine and biotechnology, fine-grained control of multiple physicochemical properties opens avenues for engineering of novel drugs and proteins, for example, antibiotics with increased efficacy and reduced toxicity \citep{wong2023discovery}, and specialized proteins with manipulated attributes like fluorescence, binding affinity \citep{shen2014introduction}, and stability \citep{chan2021deep}.

Conventional methods for multi-attribute control often rely on mechanisms such as class-conditioned LMs \citep{keskarCTRL2019, lu2022quark, hallinan-etal-2023-steer} or latent attribute embeddings \citep{He2020A, russo-etal-2020-control, riley-etal-2021-textsettr, gu-etal-2022-distributional, liu2022composable, ding2023maclasa}. However, these approaches are generally proposed to handle attributes with categorical values and may not effectively generalize to those with continuous/scalar values, such as gradually changing the complexity/readability of a sentence or modifying the activity of a protein within specified bounds.
Techniques that do support real-value constraints satisfaction often require computationally expensive gradient-based decoding optimizations \citep{Dathathri2020Plug, NEURIPS2021_79ec2a42, kumar-etal-2022-gradient, NEURIPS2022_3e25d1af, NEURIPS2022_1be5bc25} or energy-based sampling \citep{mireshghallah-etal-2022-mix, liu-etal-2023-bolt}. These methods suffer from slow inference and fixed-length outputs, limiting their widespread adaption to downstream applications. 

\begin{figure}[t]
\centering
\includegraphics[width=0.9\linewidth]{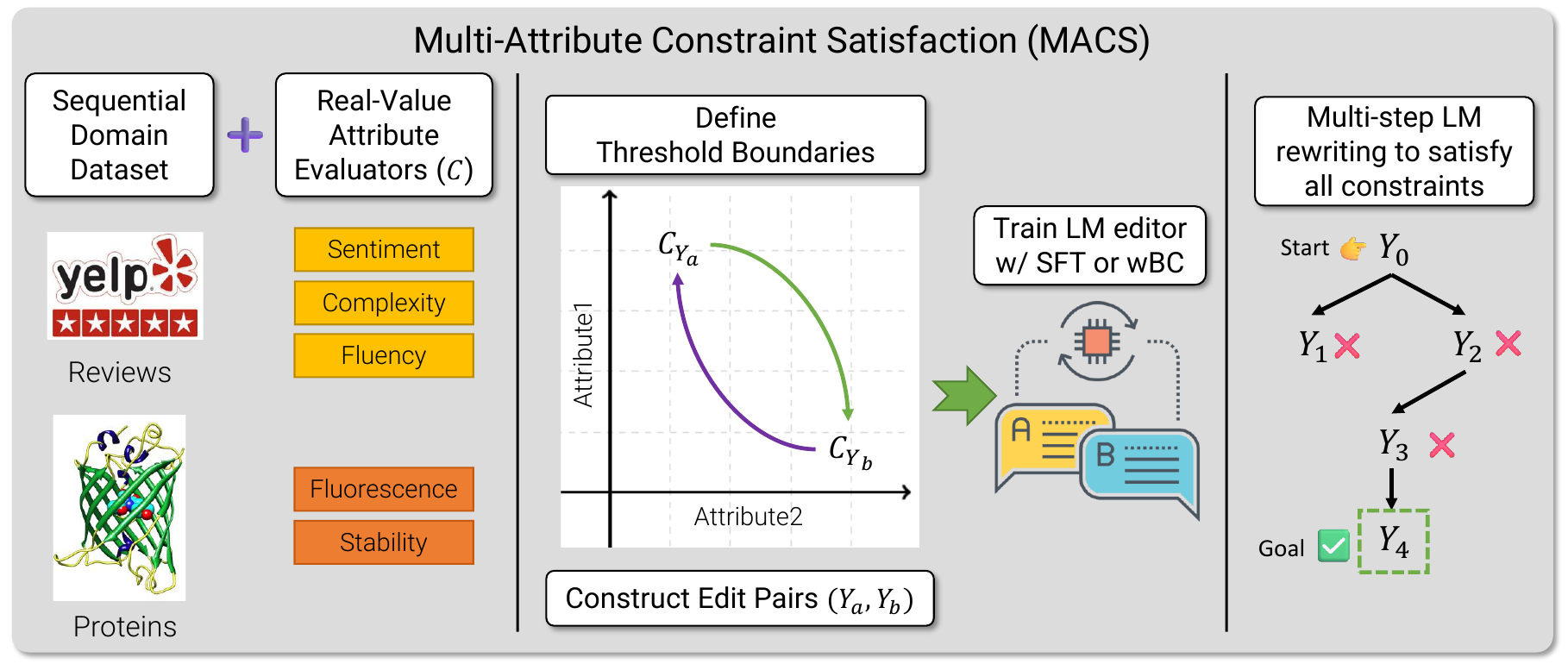}
\caption{{\framework} framework starts with sequential domain datasets (customer reviews or proteins) and a set of real-value attribute evaluators (such as sentiment, complexity regressors, or protein folding models). We then define fine-grained threshold-window boundaries for every attribute and create edit pairs distributed across the multi-attribute landscape. We train the LM editor on top of the edit pairs by leveraging supervised fine-tuning (SFT) or reward-weighted behavior cloning (wBC). Subsequently, LM editors can achieve the desired fine-grained constraints by employing prioritized editing that maintains a priority queue of past edits ordered by their proximity to the target threshold constraints.}
\label{fig:macs_overview}
\end{figure}

We introduce the \textbf{{\frameworkLong} ({\framework}) framework}, a generalized method for training LMs from diverse sequential domains towards fine-grained constraint satisfaction. Here, the LM is conceptualized as an \textbf{editor} tasked with navigating the multi-attribute landscape, iteratively refining its outputs to meet desired constraints. Unlike online and off-policy reinforcement learning (RL) methods \citep{lu2022quark, hallinan-etal-2023-steer}, which require LM editor-generated data during training, our approach utilizes an initial set of paraphrased outputs and directly trains the LM editor on them by sampling edit pairs. The initial paraphrases are obtained externally. In the case of text style transfer, for example, through few-shot prompting in the language domain. In the case of protein design, through randomized mutations in the protein domain.
As part of the framework, we introduce a generalized reward function for constraint satisfaction and experiment with offline reward-weighted behavior cloning \citep{NIPS2016_2f885d0f} to train the fine-grained LM editor on sampled edit pairs. 
Finally, we introduce a reward-prioritized multi-step inference strategy to enhance constraint satisfaction while reducing overall inference computational cost. We provide an overview of our learning and evaluation process of {\framework} in Figure \ref{fig:macs_overview}.

To comprehensively assess the effectiveness of {\framework}, we introduce a novel {\benchmarkLong} ({\benchmark}) benchmark, that consists of two challenging controllability tasks. The first task, \textit{Text Style Transfer} (\S \ref{sec:style_transfer}), requires precisely modifying the sentiment and complexity of a given text while preserving its fluency and content similarity. We sub-divide sentiment and complexity attribute ranges into five threshold window constraints each, resulting in 25 different multi-attribute threshold combinations. The second task, \textit{Protein Design} (\S \ref{sec:protein}), focuses on mutating the Green Fluorescent Protein (GFP) to achieve the desired fluorescence and stability, with both attributes divided into four threshold windows, leading to a total of 16 threshold-window combinations. During evaluation, the LM editors are tasked with satisfying every multi-attribute constraint within a fixed inference budget. We analyze the trade-off between different input conditions, inference strategies, and objective functions. 

A systematic comparison of our framework against preexisting domain-specific baselines\footnote{We only focus on offline methods which exclusively use preexisting data, thus avoid comparison with online and off-policy RL methods in our study which typically require some form of expensive LM exploration.} shows that LM editors trained with {\framework} yield the highest constraint satisfaction success rate in both {\benchmark} tasks. Our findings show the potential of adapting language models from diverse domains as fine-grained editors that allow navigating across the multiple-attribute landscape and discovering novel sequences. We release the code at \url{https://github.com/abaheti95/MACS}.

\section{Related Work}

\paragraph{Precise Multi-Attribute Control} While controlled text generation and style transfer have been widely studied problems in NLP literature, enabling fine-grained constraint satisfaction still proves to be quite challenging. A prominent approach is to incorporate attribute signals in gradients during decoding to allow satisfying multiple attribute constraints on them \citep{Dathathri2020Plug, NEURIPS2021_79ec2a42, kumar-etal-2022-gradient, NEURIPS2022_3e25d1af, NEURIPS2022_1be5bc25, liu-etal-2023-bolt}. However, there are three major limitations of these methods, (1) they require white-box access to evaluators for gradient computation, (2) their decoding speed is slow and memory intensive, and (3) their output length needs to be predefined to tractably compute the gradients. Other studies have proposed architecture augmentations and specialized loss functions \citep{russo-etal-2020-control, riley-etal-2021-textsettr, gu-etal-2022-distributional, liu2022composable, ding2023maclasa, 9944920} to perform multi-attribute control of language models. However, they typically cannot work with arbitrary external attribute evaluators and some also require expensive on-policy or off-policy samples during training. Recently, \citet{mireshghallah-etal-2022-mix} proposed probabilistic energy models to allow black-box attribute scorer constraints, but can only use masked language models for output sampling. In contrast to all the above methods, our framework leverages offline learning to offer the most flexibility in terms of external scorers, LM architecture choice, and training data sources.

\paragraph{Iterative Refinement via verbal feedback} LLMs may not generate the best output on their first attempt. Therefore, many recent prompting methods have been introduced for LLMs to iteratively improve model outputs while incorporating internal and/or external LLM evaluators as verbal feedback \citep{shinn2023reflexion, zhang-etal-2023-summit,madaan2023selfrefine, dhuliawala2023chainofverification, akyurek-etal-2023-rl4f, gou2024critic}. However, these methods implicitly expect the availability of expert large language model (LLM) which may become costly during inference. Studies also find prompting LLMs with only scalar feedback is not as effective as using both scalar and verbal feedback \citep{peng2023check}. These methods are further limited by unrecognizable language or non-language sequential data sources (for example, DNA, protein, or chemical sequences) due to lack of domain knowledge \citep{ouyang2024structured}, motivating the need for general-domain rewriting approach like {\framework}. 

\paragraph{Iterative Refinement via fine-tuning} To reduce inference costs, a few studies have demonstrated single attribute improvement across a diverse set of tasks via finetuning approaches for small LMs \citep{padmakumar2023extrapolative, welleck2023generating}. Typically, a {\em corrector}---a small LM---edits the previous response from itself or an external LLM to improve downstream task performance. These correctors are supervised finetuned on edit pairs obtained from off-policy sampling or paraphrasing techniques (mask and infill). We built upon these works to provide a unified framework for fine-grained control of multiple external attributes while only using offline data.  

\paragraph{Data-driven approaches for Protein Engineering} 
Designing proteins with desirable functionalities using limited data has been a longstanding challenge in biotechnology. 
Recent works have successfully leveraged machine learning and deep learning methods on assay-labeled data to find new protein sequences with enhanced properties such as fluorescence, binding affinity, stability, assembling, and net charge content \citep{Hsu2022, sinai2020adalead, pmlr-v162-ren22a, padmakumar2023extrapolative, kirjner2024improving, sternke2023proteinrl}. However, most of these approaches are limited to unidirectional optimization of only a single attribute that may compromise other physicochemical properties. Fine-grained control of protein sequences can allow simultaneous tuning of multiple properties of interest and provide deeper insights into sequence-structure-function relationships of proteins across these properties. For example, understanding the impact on activity \citep{HUANG1996688, Guo2004ProteinTT}, fluorescence \citep{10.1242_jcs.005801, 10.1021_ct3007452}, stability \citep{RABBANI2023822, Schlinkmann2012CriticalFF, childers2017insights}, solubility \citep{bolognesi2019mutational}, assembly \citep{garcia2022mutant, bryant2021deep} and binding affinity \citep{starr2020deep, whitehead2012optimization} under different physiological conditions.

\section{\frameworkLong}
\label{sec:LM_rewriting}
\subsection{Problem Definition}

We aim to solve multi-attribute constraint satisfaction for any sequential data as a multi-step LM rewriting task. Formally, the language model is the actor in the Markov Decision Process (MDP), that learns to navigate across a multi-attribute space defined by a set of attribute evaluators $C = \{c_1, c_2, ..., c_k\}$ (which can be classifier probability, regressor, embedding similarity, protein attribute predictors, etc). All attribute evaluators convert sequential inputs into a scalar value within a finite range ($c_j(.) \in [v_{j,min}, v_{j,max}]$). Each MDP episode begins with the initial state containing a context $x$ (that can be empty), a starting sequence $y_0$ and its attribute location $C(y_0)$ and a set of threshold window constraints $T = \{t_1, t_2, ..., t_k\}$, where $t_j = (t_{j, start}, t_{j, end})$ is the threshold boundary for attribute $c_j$. The rewriting language model $M$ iteratively edits the previous sequence until it satisfies the given threshold constraints, i.e., $P_M(y_{i+1}|x, y_i, C(y_i), T)$.\footnote{$C(y_i)$ represents a vector of attribute scores for an intermediate output $y_i$} Here, each edit $y_i \rightarrow y_{i+1}$ is considered an action, with a deterministic transition to the next state. During inference, the goal is to generate a series of consecutive edits starting from $y_0$ to $y_n$, such that $C(y_n) \in T$. 


\subsection{{\framework} Approach}
\label{subsec:macs_approach}

\paragraph{Edit Pairs Construction} Even though the rewriting process is inherently multi-step during inference, we can isolate individual edits and train language model rewrite using offline pairs. For example, given any pair of similar sequences $y_a$ and $y_b$ which have distinct attribute locations $C(y_a)$ and $C(y_b)$, we can construct a training instance by asking the language model to edit $y_a \rightarrow y_b$ and artificially selecting threshold windows $T_{a\rightarrow b}$\footnote{Selecting the threshold window that satisfies $C(y_b)$ works best for training the language model editor.} that encourage $M$ to move from $C(y_a)$ towards $C(y_b)$ \cite{NIPS2017_453fadbd}. We can similarly define another training instance going from $y_b \rightarrow y_a$. Assuming $m$ variations of a particular sequence are available ($y_1, y_2, ... y_m$), we can construct $P^m_2$ trainable \textit{edit pairs} from them. In \S \ref{sec:style_transfer} and \S \ref{sec:protein} we show how we create edit pairs for languages and proteins respectively.

\paragraph{Constraint Satisfaction Reward} We want to encourage the rewriter LM to make edits that move closer to the user-provided multi-attribute threshold boundary. If the initial sequence is already inside the target threshold boundaries, we expect the LM to paraphrase the sequence. Based on these two aspects, for each attribute ($c_j(.) \in [v_{j,min}, v_{j,max}]$) and its corresponding threshold boundary ($t_j = (t_{j, start}, t_{j, end})$), we define its constraint satisfaction reward as the sum of the two components,
\begin{equation}
    R(y_n, y_o, c_j(.), t_j) = \underbrace{f(c_j(y_n), t_j)}_{\text{Satisfaction Score}} + \underbrace{f(c_j(y_n), t_j) - f(c_j(y_o), t_j)}_{\text{Change in Satisfaction Score}}
\end{equation}
Here $y_n$ and $y_o$ represent the new and the old sequence respectively, while $f(.) \in [0,1]$ is the threshold satisfaction scoring function that shows the deviation of the attribute score from its threshold boundary. We set the satisfaction score as $1$ if its attribute location satisfies the threshold and linearly decreases to $0$ as it moves towards the extreme ends,
\begin{equation}
f(c_j(y), t_j) = \begin{cases}\frac{(c_j(y) - v_{j,min})}{(t_{j, start} - v_{j,min})} & \text{if } c_j(y) < t_{j, start} \\
     1 & \text{if } t_{j, start} \le c_j(y) \le t_{j, end} \\
    \frac{(v_{j,max} - c_j(y))}{(v_{j,max} - t_{j, end})} & \text{otherwise}
    \end{cases}
\end{equation}

The total multi-attribute reward is defined as the sum of satisfaction reward for all attributes,
\begin{equation}
\label{eq:reward}
    R(y_n, y_o, C, T) = \sum_j^k R(y_n, y_o, c_j(.), t_j)
\end{equation}

\paragraph{Learning} Given a collection of edit pairs $D = \bigcup_{x, y_a, y_b} \{(x, y_a, y_b, T_{a\rightarrow b})\}$ we obtain an LM editor by employing supervised finetuning, e.g., with the negative log-likelihood loss $\mathcal{L}_{SFT}(M) = - \ln P_M(y_b|x, y_a, C(y_a), T_{a\rightarrow b})$. To improve beyond the supervised finetuned model, we experiment fine-tuning it further with the offline reward-weighted behavior cloning objective \citep{NIPS2016_2f885d0f, junczys-dowmunt-etal-2018-approaching, NEURIPS2020_588cb956, ghosh-etal-2021-helpful, ramachandran-etal-2022-caspi, yang2023preferencegrounded, feng2023fantastic, baheti2024leftover}, that directly multiplies the reward with SFT objective $\mathcal{L}_{wBC}(M) = -  R(y_b, y_a, C, T_{a\rightarrow b}) \times \ln P_M(y_b|x, y_a, C(y_a), T_{a\rightarrow b})$. We provide the pseudo-code of the {\framework} training process in Algorithm \ref{alg:main}.

\setlength{\algomargin}{10pt}
\begin{algorithm}[t]
\caption{{\frameworkLong} Training pseudo code}\label{alg:main}
\KwData{Edit Pairs Offline set $D = \bigcup_{x, y_a, y_b} \{(x, y_a, y_b, T_{a\rightarrow b})\}$, Attribute Evaluators $C = \{c_1, c_2, ..., c_k\}$, Initial rewriting language model $M$, SFT steps $N_1$, wBC steps $N_2$, Learning rates $\alpha_1, \alpha_2$}
\nl Obtain attribute values for all sequences $y_i \in D$ \\
\nl $M_1 \leftarrow M$ \\
\nl \For{$i \leftarrow 1$ \KwTo $N_1$}{
    \nl Sample edit pair $(x, y_a, y_b, T_{a\rightarrow b})$ from $D$ (random or k-NN sampling) \\
    \nl $\mathcal{L}_{SFT}(M_1) = - \ln P_{M_1}(y_b|x, y_a, C(y_a), T_{a\rightarrow b})$ \\
    \nl $M_1 \leftarrow M_1 - \alpha_1 \nabla_{M_1} \mathcal{L}_{SFT}(M_1)$ \\
}
\nl $M_2 \leftarrow M_1$ \\
\nl \For{$i \leftarrow 1$ \KwTo $N_2$}{
    \nl Sample edit pair $(x, y_a, y_b, T_{a\rightarrow b})$ from $D$ (random or k-NN sampling) \\
    \nl $\mathcal{L}_{wBC}(M_2) = -R(y_b, y_a, C, T_{a\rightarrow b}) \times \ln P_{M_2}(y_b|x, y_a, C(y_a), T_{a\rightarrow b})$ \\
    \nl $M_2 \leftarrow M_2 - \alpha_2 \nabla_{M_2} \mathcal{L}_{wBC}(M_2)$ \\
}
\nl Evaluate $M_1$ and $M_2$ with multi-step inference strategies\\
\end{algorithm}

\paragraph{Multi-Step Reward Prioritized Inference} Satisfying multiple precise constraints $T$ across diverse attributes $C$ is a challenging task for which the editor language model, $P_M(y_{i+1}|x, y_i, C(y_i), T)$, may not get the correct answer in one try. A trivial solution is to employ a best-of-N inference strategy. To improve beyond best-of-N, previous solutions to single attribute iterative improvement propose using an iterative editing strategy, where the rewriter LM generates a trajectory of edits sequentially ($y_0 \rightarrow y_1 ... \rightarrow y_n$) \citep{padmakumar2023extrapolative, welleck2023generating}. However, this naive editing strategy doesn't interact with the attribute evaluators and cannot verify if the intermediate edits are moving toward the threshold constraints or not. We instead propose maintaining a priority queue of edits using the generalized reward function we defined in equation \ref{eq:reward}. In this strategy, the LM generated subsequent edit $y_i \rightarrow y_{i+1}$ is only retained if it moves closer to the threshold satisfaction, i.e. $R(y_{i+1}, y_0, C, T) > R(y_i, y_0, C, T)$. We call this strategy \textit{prioritized} inference and compare its performance against best-of-N and naive editing.

In the subsequent sections, we extensively and systematically evaluate {\framework} and baselines on a new {\benchmarkLong} (\benchmark) benchmark, which comprises two fine-grained control tasks: Text Style Transfer \S \ref{sec:style_transfer} and Protein Design \S \ref{sec:protein}.


\section{{\benchmark} - Text Style Transfer}
\label{sec:style_transfer}

While foundational large language models are capable of solving a variety of general language tasks via prompt engineering \citep{ouyang2022training, openai2024gpt4}, they incur large computation overhead during inference and often underperform in directly incorporating external real-value feedback \cite{peng2023check}. To mitigate their limitations, we develop {\framework} to fine-tune small language models that enable constraint satisfaction on external signals via iterative refinement  \citep{padmakumar2023extrapolative, welleck2023generating}. To evaluate these methods, we create {\benchmark} - Text Style Transfer task, where the goal is to precisely modify the sentiment and complexity of Yelp reviews while preserving fluency and content similarity. 

\paragraph{Attribute Evaluators} To obtain the sentiment and complexity evaluators, we train RoBERTa-large \citep{liu2020roberta} regressors on Yelp reviews \citep{NIPS2015_250cf8b5} and the SWiPE Wikipedia simplification dataset \citep{laban-etal-2023-swipe}. The output range of the sentiment regressor is $\in [1,5]$, while the complexity regressor is within the range $\in [-2, 2]$.\footnote{Training details of sentiment and complexity regressors is provided in Appendix \ref{sec:sentiment_complexity_regressor}} Subsequently, we defined five threshold boundaries for sentiment as follows: $(1, 1.5)$ very negative, $(1.5, 2.5)$ negative, $(2.5, 3.5)$ neutral, $(3.5, 4.5)$ positive, $(4.5, 5)$ very positive and five threshold boundaries for complexity as follows, $(-2, -1.5)$ very simple, $(-1.5, -0.5)$ simple, $(-0.5, 0.5)$ normal, $(0.5, 1.5)$ complex, $(1.5, 2)$ very complex. In total, these results in 25 different multi-attribute threshold combinations.

We further include two more evaluators to encourage fluency and content preservation: (1) fluency classifier probability ($\in [0,1]$)\footnote{RoBERTa-base classifier from the Corpus of Linguistic Acceptability \citep{warstadt-etal-2019-neural} \href{https://huggingface.co/textattack/roberta-base-CoLA}{textattack/roberta-base-CoLA}} and cosine text embedding similarity score\footnote{\href{https://huggingface.co/sentence-transformers/all-mpnet-base-v2}{sentence-transformers/all-mpnet-base-v2}} between the previous and the new output ($\in [0,1]$). Since we always want to maximize both properties, we add their scores directly in the constraint satisfaction reward function (eqn. \ref{eq:reward}) as two additional components.

\begin{figure}
\begin{subfigure}[t]{0.45\textwidth}
  \centering
  \includegraphics[width=\linewidth]{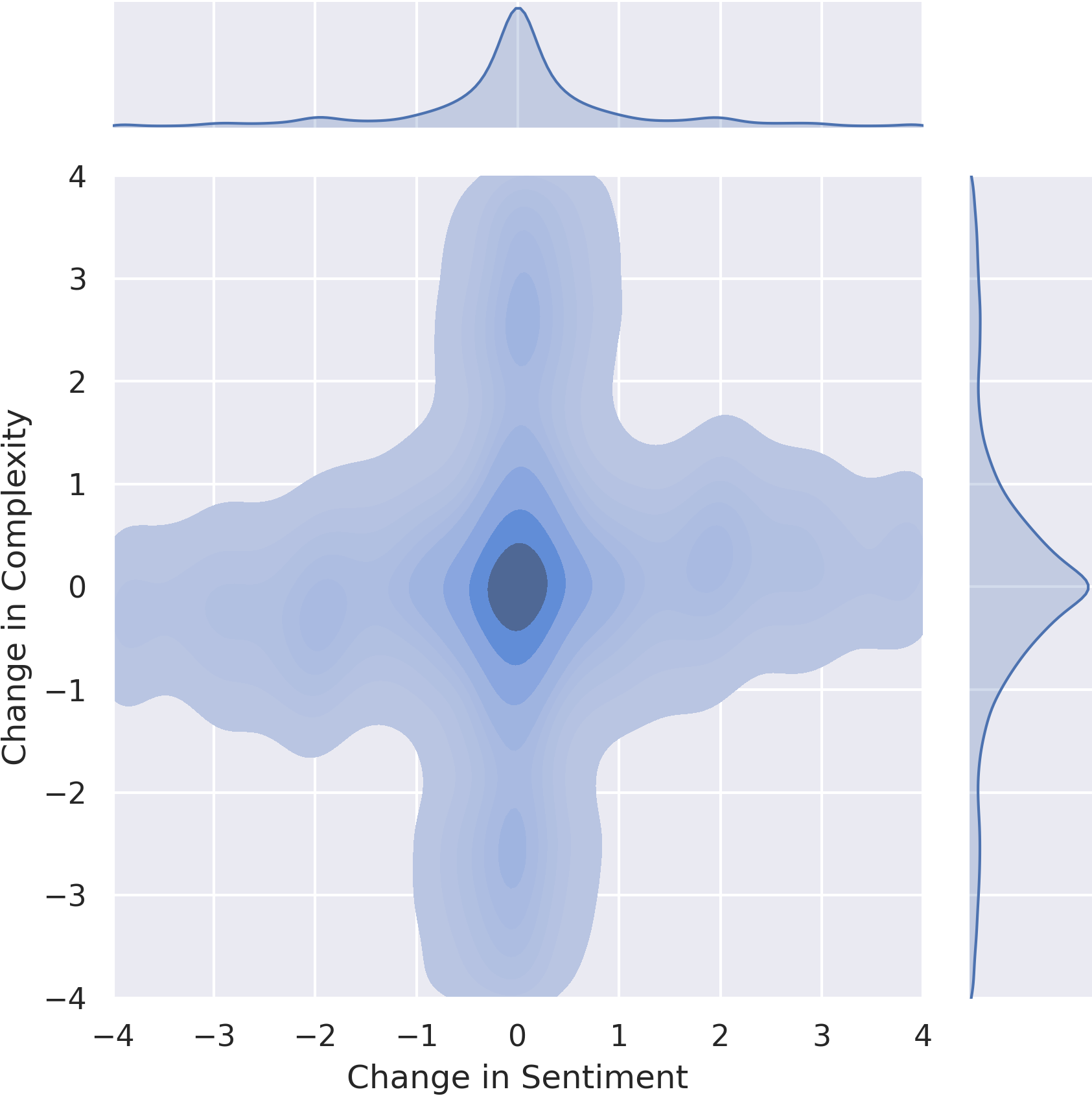}
  \caption{Random Sample Density} 
  \label{fig:random_density}
\end{subfigure}\hfill
\begin{subfigure}[t]{0.45\textwidth}
  \centering
  \includegraphics[width=\linewidth]{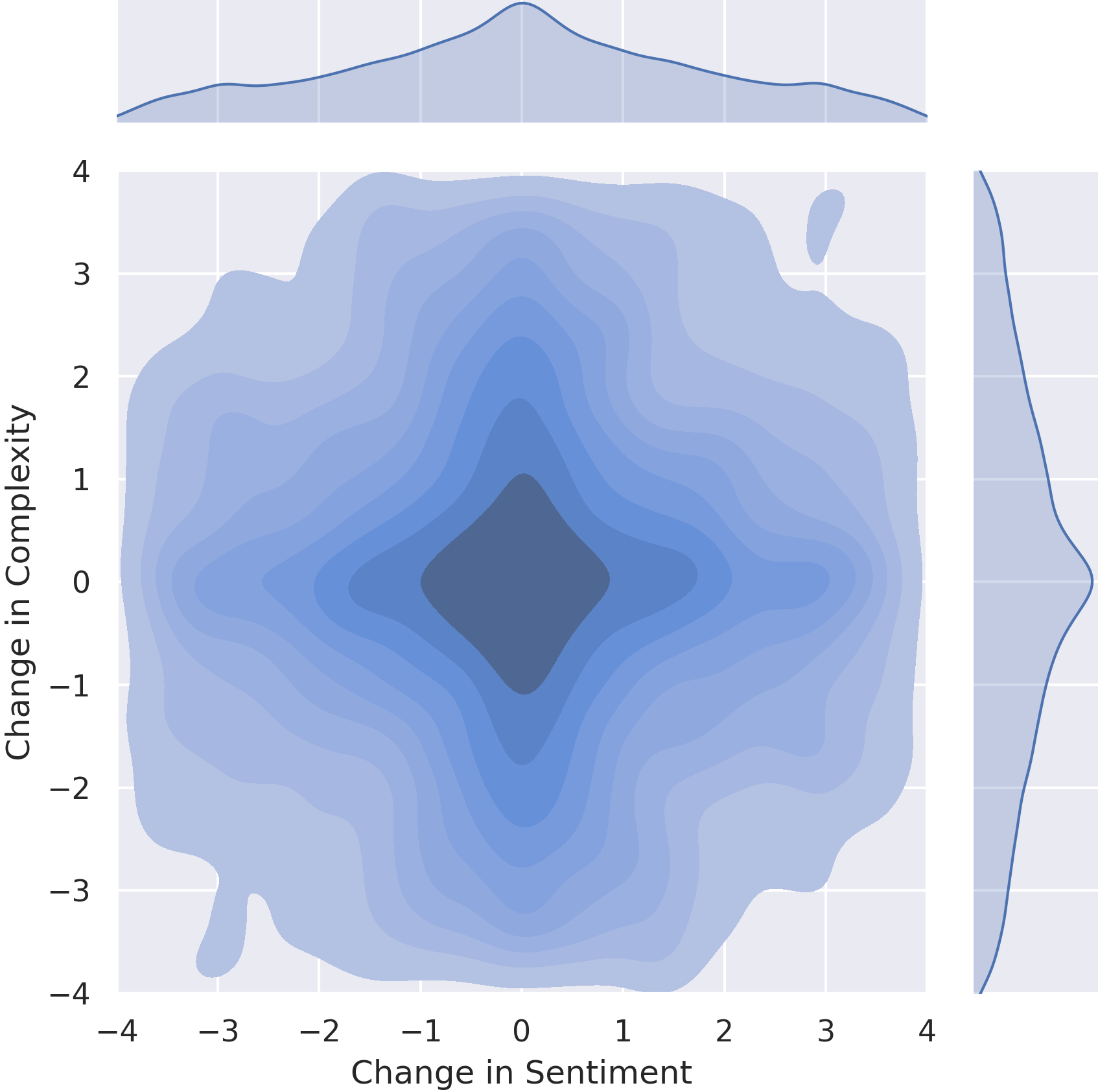}
  \caption{k-NN Sample Density} 
  \label{fig:knn_density}
\end{subfigure}
\caption{Sentiment and Complexity attribute edit pair distribution via random sampling vs. proposed k-NN sampling. k-NN sampling yields a much more diverse set of edit pairs better suited for simulating editing in all directions.}
\label{fig:random_dist}
\end{figure}

\paragraph{Creating Attributed Variations and Edit Pairs} To synthetically obtain a diverse set of paraphrases previous studies have proposed various techniques such as mask-then-infill \citep{xu-etal-2018-unpaired, li-etal-2018-delete, ma2020powerTransformer, padmakumar2023extrapolative}, back-translations \citep{prabhumoye-etal-2018-style, zhang2018style, lample2018multipleattribute, Luo19DualRL}, paraphrasing models \citep{krishna-etal-2020-reformulating} and generating multiple samples \citep{welleck2023generating}. In our preliminary experiments, these methods did not yield many diverse attribute variations. Instead, we use few-shot prompted LLMs to generate alternate attributed variations of reviews. In particular, for both sentiment and complexity attributes, we first sample an equal number of reviews from each threshold boundary ($1K$ from each label, $5K$ total for each attribute). Then, we construct few-shot LLM prompts that propose five alternate variations of each review, one for each sentiment (or complexity) label. We employ nucleus sampling \citep{holtzman2019curious} on the few-shot prompts ($top_p=0.95$) to generate $25$ variations for each review, $\approx 125K$ total variations for each attribute.\footnote{The LLM-generated variations do not always agree with the target thresholds but provide a good spread of paraphrases in the multi-attribute space. We filter out variations that yield fluency score or embedding similarity score $<0.7$.} Considering the original review and its $25$ new proposed variations, we can construct at most $P^{26}_2$ trainable edit pairs for each review (\S \ref{subsec:macs_approach}). The final dataset simply combines all the edit pairs from sentiment and complexity variations. We choose Llama2-7B parameter model \citep{touvron2023llama} as our base LLM and provide the prompts designed for sentiment and complexity attributes in the Appendix \ref{sec:few_shot_variations}. 

Within the language domain, edit pairs from synthetic variations are not uniformly distributed. Therefore we propose \textit{k-NN Sampling} to obtain evenly distributed edit pairs in the multi-attribute space. In multi-step editing via LM, consecutive edits can lead to a large drift in content from the original text. Subsequently, we propose an \textit{Anchor Conditioned Inference} strategy to mitigate this problem. We discuss both algorithmic modifications below. 


\paragraph{k-NN Edit Pair Sampling} An ideal data distribution should have edit pairs from every multi-attribute location to every other location. However, in practice, this is not always true. Given an edit pair $y_a \rightarrow y_b$, we visualize its attribute change by converting the difference into vector $C(y_b) - C(y_a)$. We show the distribution of edit pairs when sampled randomly from our synthetic variations in Figure \ref{fig:random_density}. It shows that most of the edits change only one attribute on average. To obtain a more balanced coverage, we propose using a k-nearest neighbors (k-NN) sampling strategy as follows, (1) sample two multi-attribute threshold boundaries at random, (2) uniformly sample attribute locations from both boundaries (representing start and end) (3) find k-nearest neighbors ($k=30$) of the sampled transition from the available edit pairs and randomly select one of them. We find the edit pairs sampled with the k-NN strategy to be much more evenly distributed (Figure \ref{fig:knn_density}).

\paragraph{{\faAnchor} Anchor Conditioned Inference} To reduce content drift in multi-step rewriting, we propose to include the original text in the context of the LM's prompt which we call \textit{anchor conditioning}. During multi-step inference, when a new output sequence is generated, we still retain the original text in the context of subsequent rewrites. To train the rewriter LM with anchor, we augment $D$'s edit pairs $(y_a \rightarrow y_b)$ by sampling an anchor $y_c$ such that $R(y_c, y_a, C, T_{a\rightarrow b}) \ge R(y_b, y_a, C, T_{a\rightarrow b})$. In the experiments, we evaluate the effectiveness of this anchor conditioning in content preservation over multiple rewrites. 

\subsection{Text Style Transfer Evaluation}

For the Text Style Transfer task on Yelp Reviews, we design a fixed inference budget evaluation setup, i.e., each method will have a fixed number of allowed rewrites to satisfy all multi-attribute constraints. Subsequently, we construct a test set of $250$ total reviews ($10$ from each of the $25$ sentiment and complexity threshold combinations). The task is to generate $25$ attributed paraphrases for every test review within $5$ rewrites ($250 \times 25 \times 5 \approx 31.2K$ total inference budget). For every baseline and our models, we compare the constraint satisfaction success rate of multi-step inference strategies: best-of-N, naive rewriting, and reward-prioritized rewriting. We report the average satisfaction rate, fluency, and embedding similarity of paraphrases that satisfied the given constraints.

\paragraph{Baselines and {\framework} models}
As a baseline, we use few-shot prompted Llama2-7B \citep{touvron2023llama} and Llama3-8B \citep{llama3modelcard} models as fine-grained editors for our Text Style Transfer task. For every transition from one threshold combination to another, we find 10 edit pairs as few-shot demonstrations (a total of $25 \times 25 \times 10 = 6250$ edit pairs). For fine-tuning methods, we use a smaller TinyLlama \citep{zhang2024tinyllama} 1.1B parameter model as the multi-attribute rewriter LM. Among finetuning baselines, we compare with Control Tokens \citep{keskarCTRL2019} that simply convert each threshold combination into style tokens. We allocate 10 total style tokens (5 for each attribute) and simply append the style tokens of the target threshold windows in the prompt along with the target response as follows: $y_a$\texttt{[Sentiment Token][Complexity Token]}$y_b$. For LMs trained with {\framework}, we construct a \textit{text-only} prompt that doesn't use any special tokens as follows: 

\begin{allintypewriter}
Review: $y_a$\\
Review's Sentiment: $c_1(y_a)$ and Complexity: $c_2(y_a)$

Paraphrase the review such that its Sentiment is within: $t_{1, a\rightarrow b}$ and Complexity is within: $t_{2, a\rightarrow b}$\\
Paraphrased Review: $y_b$
\end{allintypewriter}
We prepend the above text prompt with $y_c$ and its attribute locations for anchor conditioning ({\faAnchor}) training.

We train both control tokens and text-prompted models with supervised fine-tuning (SFT) for 200K steps and a batch size of 16. For control tokens, we experiment with both randomized and k-NN edit pair sampling, whereas we only use k-NN edit pair sampling for text-based models. For weighted behavior cloning (wBC) objective, we continue training the supervised finetuned models for an additional 50\% steps (100K steps).

\begin{table}[t]
\centering
\small
\caption{{\benchmark} Text Style Transfer task evaluation: Paraphrase each test review 25 times into fine-grained Sentiment and Complexity threshold constraints while maintaining fluency and content preservation. We compare the Control Tokens baseline with supervised fine-tuned and wBC models each with 3 different inference strategies: Best-of-N, naive rewriting, and reward-prioritized rewriting. We report the average satisfaction rate for each model and inference strategy and average fluency and embedding similarity of the paraphrases that satisfied the constraints. \textbf{\textit{Takeaway}:} Our proposed reward-prioritized rewriting combined with anchor conditioning ({\faAnchor}) and wBC obtains the highest satisfaction rate. However, its differences with ({\faAnchor}) and SFT are not statistically significant\textsuperscript{\ding{61}}, indicating that anchor conditioning leads to most improvement in multi-step editing.}
\label{tab:st_main_result}
\resizebox{0.98\textwidth}{!}{
\begin{tabular}{c|l|c|ccc|ccc|ccc}
\toprule
& \multicolumn{2}{c}{Inference Type} & \multicolumn{3}{|c}{Best-of-N} & \multicolumn{3}{|c}{Naive Rewriting} & \multicolumn{3}{|c}{Reward-Prioritized} \\
\cline{2-12}
& \multirow{2}{*}{Method} & \multirow{2}{*}{\parbox{1.1cm}{\centering Train Sample}} & \multirow{2}{*}{\parbox{1.4cm}{\centering Satisfaction Rate*}}  & \multirow{2}{*}{\parbox{0.5cm}{\centering Flue-ncy}} & \multirow{2}{*}{\parbox{0.5cm}{\centering Emb. Sim.}} & \multirow{2}{*}{\parbox{1.4cm}{\centering Satisfaction Rate*}}  & \multirow{2}{*}{\parbox{0.5cm}{\centering Flue-ncy}} & \multirow{2}{*}{\parbox{0.5cm}{\centering Emb. Sim.}} & \multirow{2}{*}{\parbox{1.4cm}{\centering Satisfaction Rate*}} & \multirow{2}{*}{\parbox{0.5cm}{\centering Flue-ncy}} & \multirow{2}{*}{\parbox{0.5cm}{\centering Emb. Sim.}} \\
& & & & & & & & & & \\
\midrule
\parbox[t]{2mm}{\multirow{4}{*}{\rotatebox[origin=c]{90}{baselines}}} & \multicolumn{2}{l|}{10-shot Llama2-7B} & $.478\pm .141$ & $.93$ & $.85$ & - & - & - & - & - & - \\
& \multicolumn{2}{l|}{10-shot Llama3-8B} & $.594\pm .130$ & $.93$ & $.85$ & - & - & - & - & - & - \\
\cline{2-12}
& Control Tokens & random & $.774\pm.083$ & $.92$ & $.80$ & $.783\pm.068$ & $.92$ & $.78$ & $.792\pm.061$ & $.93$ & $.79$ \\
& Control Tokens & k-NN & $.828\pm.063$ & $.92$ & $.80$ & $.809\pm.046$ & $.91$ & $.78$ & $.828\pm.048$ & $.92$ & $.79$ \\
\midrule
\parbox[t]{2mm}{\multirow{4}{*}{\rotatebox[origin=c]{90}{\framework}}} & SFT & k-NN & $.824\pm.061$ & $.92$ & $.80$ & $.809\pm.056$ & $.92$ & $.78$ & $.827\pm.054$ & $.93$ & $.79$ \\
& SFT + wBC & k-NN & $.820\pm.072$ & $.92$ & $.81$ & $.815\pm.063$ & $.92$ & $.79$ & $.835\pm.051$ & $.93$ & $.80$ \\
\cline{2-12}
& {\faAnchor} + SFT & k-NN & $.833\pm.065$ & $.92$ & $.81$ & $\mathbf{.849}\pm.054$\textsuperscript{\ding{61}} & $.92$ & $.80$ & $.847\pm.052$\textsuperscript{\ding{61}}  & $.92$ & $.80$ \\
& {\faAnchor} + SFT + wBC & k-NN & $\mathbf{.835}\pm.065$ & $.92$ & $.81$ & $.840\pm.061$ & $.92$ & $.80$ & $\mathbf{.855}\pm.059$\textsuperscript{\ding{61}}  & $.92$ & $.80$ \\
\bottomrule
\end{tabular}
}
\end{table}

\begin{figure}[t]
\begin{subfigure}[t]{0.49\textwidth}
  \centering
  \includegraphics[width=\linewidth]{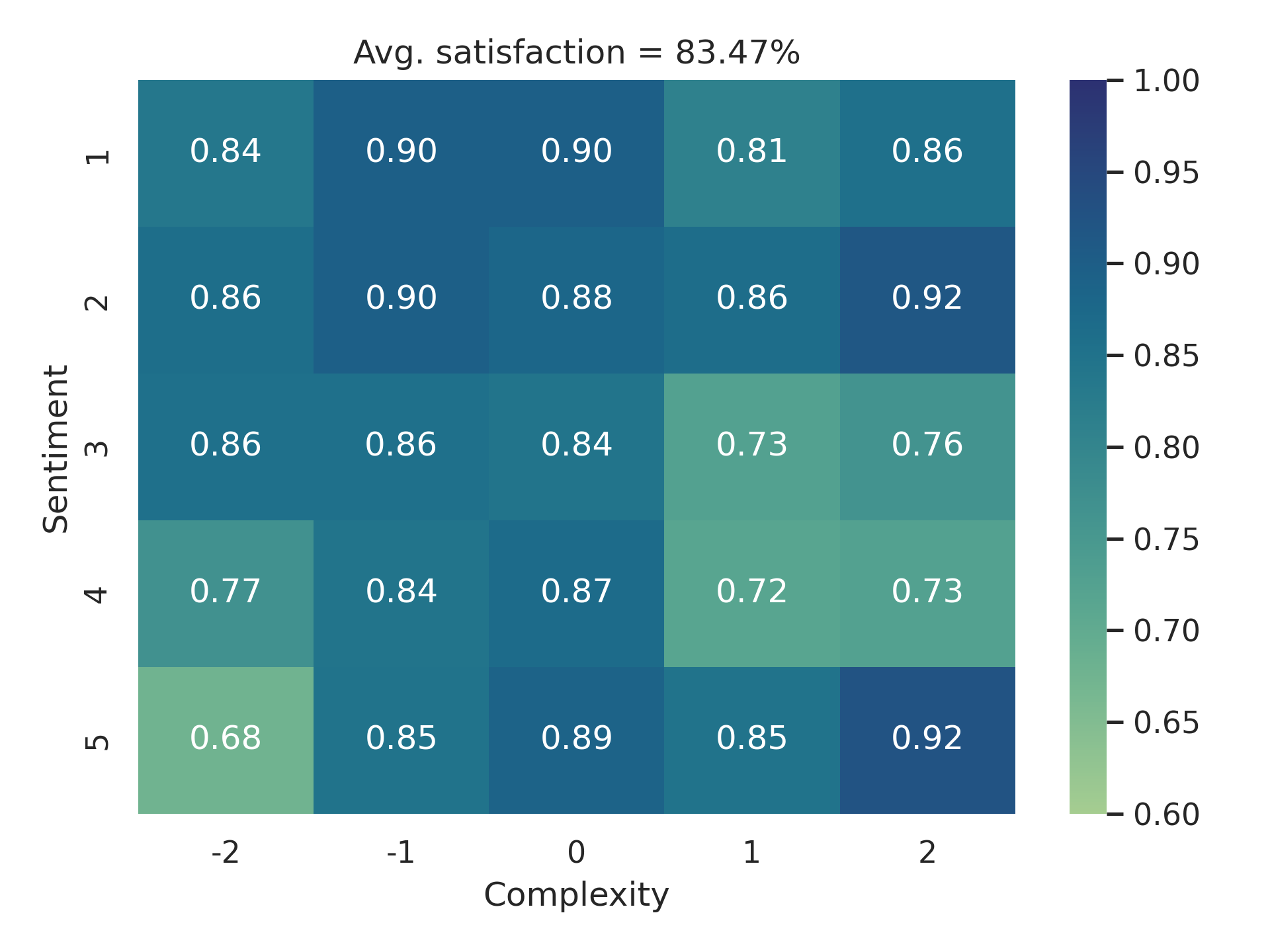}
  \caption{{\faAnchor} + SFT + wBC Best-of-N inference} 
\end{subfigure}\hfill
\begin{subfigure}[t]{0.49\textwidth}
  \centering
  \includegraphics[width=\linewidth]{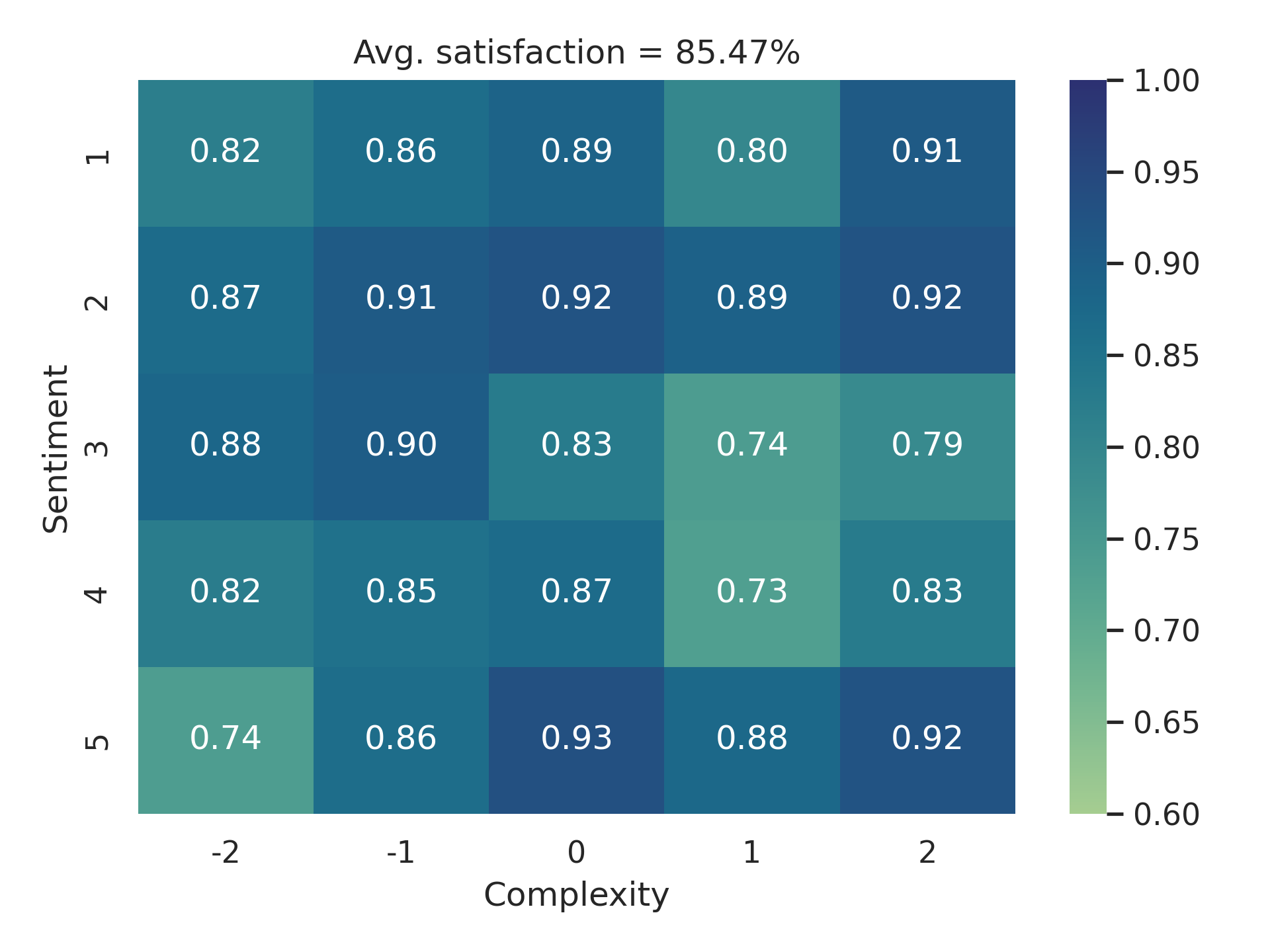}
  \caption{{\faAnchor} + SFT + wBC Reward-Prioritized inference} 
\end{subfigure}
\caption{Comparing best-of-N vs. reward-prioritized inference constraint satisfaction rate of Sentiment and Complexity attributes. \textbf{\textit{Takeaway}:} Reward-prioritized inference has better satisfaction rates in \textit{harder to reach} constraints i.e. edges of the satisfaction matrix.}
\label{fig:st_statisfaction}
\end{figure}

\subsection{Text Style Transfer Results}
We present the performance of all baselines and {\framework} models with the three inference types in Table \ref{tab:st_main_result}. Among few-shot methods, the newer Llama3 model outperforms Llama2, however, both struggle to achieve very high satisfaction rates and show high variance across different threshold combinations. In comparison, the control tokens-based finetuning baseline works much better than few-shot prompting and gains a further boost in overall satisfaction rate when trained with our proposed k-NN edit pair sampling. Interestingly, naive rewriting is occasionally worse than best-of-N inference, indicating that models may not consistently move toward the threshold boundaries. The reward-prioritized rewriting improves over naive rewriting by leveraging the external scorers and our reward function to guide its search process.

\begin{figure}[t]
\centering
\includegraphics[width=\linewidth]{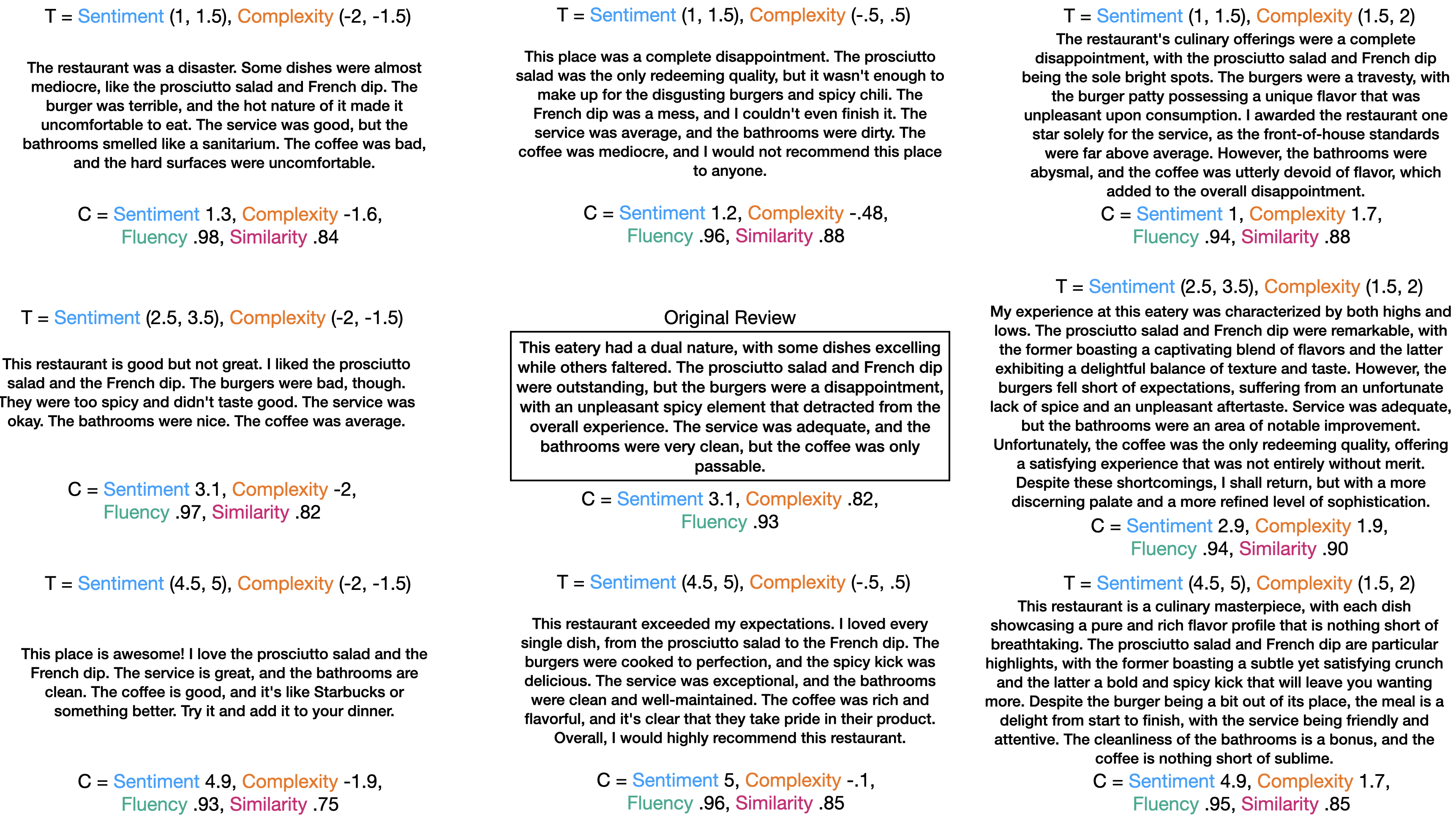}
\caption{Showing 8 attributed paraphrases of a test review for various thresholds generated by {\faAnchor} + SFT + wBC model with reward-prioritized rewriting.}
\label{fig:style_transfer_example}
\end{figure}

Among our methods, the \textit{text-only} finetuned model matches the performance of the control tokens baseline when trained with the proposed k-NN edit pair sampling.  For models without anchor conditioning, we notice that multi-step rewriting can drift away from the original content indicated by a drop in embedding similarity when switching from best-of-N to rewriting. Anchor conditioning ({\faAnchor}) resolves the content drift problem and subsequently improves satisfaction rate and final embedding similarity when employing rewriting inference strategies. Finally, we notice that models trained with wBC outperform their counterpart SFT-only models in reward-prioritized rewriting. However, the performance differences are not statistically significant ($p\approx0.2$) according to the two-proportions z-test \cite{fleiss2013statistical}. We show the detailed threshold satisfaction matrix of best-of-N vs. reward-prioritized inference for our wBC model in Figure \ref{fig:st_statisfaction}. Reward-prioritized inference achieved a better satisfaction rate than best-of-N in most of the threshold constraints, especially for harder-to-reach threshold constraints (corners of the threshold satisfaction matrix). We also present example generations from the best method in Figure \ref{fig:style_transfer_example}, showcasing the difficulty of the task. 

\section{{\benchmark} - Protein Design}
\label{sec:protein}

Unlike language, where text can be paraphrased in many different ways, fine-grained editing in protein space is challenging due to (a) the uneven distribution of assay-labeled data across multiple attributes and (b) the existence of limited potential solutions in nature for a given set of attribute constraints \citep{sternke2023proteinrl}. Moreover, for any given set of constraints, obtaining multiple novel and diverse candidates is important to maximize the chances of success in wet lab experiments \citep{pmlr-v162-jain22a}. To evaluate fine-grained control of {\framework} framework in protein space, we create {\benchmark} - Protein Design, where the task is to simultaneously modulate fluorescence and folding stability of Green Fluorescent Protein (GFP), (a protein widely investigated and used as biosensors in life sciences research).

\paragraph{Fluorescence and folding Stability Evaluators}

We obtain the dataset of $\approx 51.7K$ mutants of the GFP \textit{wild-type} (i.e., the protein sequence that occurs in nature) \citep{sarkisyan2016local, 10.7554/eLife.75842}. The dataset contains fluorescence levels on logarithm 10 scale for every mutant sequence $\in [1.28, 4.12]$. Due to a lack of assay-labeled data for a second attribute, we calculate the theoretical folding stability values ($\Delta\Delta G$ or ddG) of each mutant with respect to the wild-type structure using FoldX software\citep{schymkowitz2005foldx}.\footnote{Foldx uses an empirical force field to determine the effect of mutations on the protein folding. We note that FoldX-generated values are not an accurate representation of real experimental folding stability and are only used as a proxy. We follow best practices recommended in the previous research and compute ddG on an average of five FoldX calculations \citep{chan2021deep}.} The wild-type ddG is $0$ and any 
mutant with negative ddG is more stable than wild-type. The overall distribution of ddG for all the mutants is $\in [-5.66, 60.75]$. We train ESM2-based regressors \citep{lin2023evolutionary} as evaluators for both attributes using mean squared error loss.\footnote{We divide the dataset into 50\% train, 15\% validation, and 35\% test set for both fluorescence and ddG attributes.} The test set correlation for fluorescence and ddG are $0.974$ and $0.987$ respectively.\footnote{Implementation details in Appendix \ref{subsec:protein_regressor_impl}}

\paragraph{Attribute Distribution and Edit Pairs} We plot the distribution of log fluorescence and ddG of all the GFP mutations in Figure \ref{fig:GFP_dist} in the Appendix. Unlike language data, protein mutants are even more unevenly distributed across the multi-attribute landscape, with the bulk of the mutants clustered near the wild-type (WT) GFP sequence (which has $\approx 3.72$ log fluorescence and $0$ ddG). To effectively navigate this skewed distribution, we define four threshold boundaries in log fluorescence, $(<3.0)$ - very low, $(3.0, 3.4)$ - low, $(3.4, 3.7)$ - medium, $(> 3.7)$ - bright and four threshold boundaries in ddG, $(<0)$ - more stable than WT, $(0.0, 0.5)$ - as stable as WT, $(0.5, 2.0)$ - slightly destabilized, $(> 2.0)$ - highly destabilized \citep{doi:10.1146/annurev.biophys.37.092707.153558}. 

The limited viable solutions in certain regions ($<10\%$ of proteins have $<0$ ddG) make the protein design task very challenging, especially when learning from an offline dataset of mutations. Here, all GFP mutants are considered \textit{paraphrases} of each other, and thus, total possible edit pairs are $\approx P^{51.7K}_2$. To train the LM rewriting models to edit in all possible directions, we employ the following edit pair sampling strategy: (1) pick two multi-attribute threshold boundaries, (2) sample a mutant at random from both of the selected threshold constraints and (3) construct an edit pair by treating the first as the source and the second as the target mutant.



\subsection{Protein Design Evaluation}
Unlike the Style Transfer task, where we only care about one solution for each constraint, the goal of the Protein Design task is to find the maximum number of new mutants in every multi-attribute constraint under a fixed inference budget. For each threshold constraint, we initiate multiple random walks of different lengths starting from wild-type GFP sequence ($WT \rightarrow y_1 ... \rightarrow y_n$). We assign a total 3000 inference budget which results in (1) 3000 $\times$ 1-hops, (2) 1000 $\times$ 3-hops, and (3) 300 $\times$ 10-hops random walks. We expect duplicated predictions under specific constraints since certain regions will have naturally very few solutions. Among the 3000 predictions in each inference method, we calculate the \textit{total success rate}: ratio of distinct mutants that satisfy the constraints according to our evaluators and \textit{unique success rate}: ratio of unique successful mutants outside of the offline training data. We also compare with reward-prioritized walks from wild-type (\S \ref{sec:LM_rewriting}) where the LM generated intermediate edit $y_i \rightarrow y_{i+1}$ is only retained if it moves closer to the threshold constraints, i.e. $R(y_{i+1}, WT, C, T) > R(y_i, WT, C, T)$. We experiment with reward-prioritized walks in 1000 $\times$ 3-hops and 300 $\times$ 10-hops settings. In total, for 16 multi-attribute threshold constraints of log fluorescence and ddG, we have a total budget of $16\times 3K = 48K$ decoding in every inference method.

\begin{table}[t]
\centering
\small
\caption{{\benchmark} Protein Design task evaluation: Starting from GFP wild-type, discover the maximum possible unique mutants across 16 multi-attribute constraints of log fluorescence and ddG within 3000 total inferences. We compare the ProtGPT2 LM editor fine-tuned with SFT and wBC with 5 different inference strategies: three random walks and two reward-prioritized walks with different hop lengths. After discarding all duplicate solutions, we report the average rate of mutants that satisfy the threshold constraints (total success rate) and the average rate of successful mutants that are outside the training data (unique success rate). We also report the average edit distances between all pairs of successful mutants for each method. \textbf{\textit{Takeaway}:} wBC $\backslash$w entropy, another variant of {\framework} method, discovers the most number of novel mutants. However, the differences between different inference methods are not statistically significant\textsuperscript{\ding{61}}.}
\label{tab:pd_main_result}
\resizebox{0.98\textwidth}{!}{
\begin{tabular}{l|ccc|ccc|ccc}
\toprule

  & \multirow{3}{*}{\parbox{1.1cm}{\centering Total Success Rate}} & \multirow{3}{*}{\parbox{1.1cm}{\centering Unique Success Rate*}} & \multirow{3}{*}{\parbox{1.1cm}{\centering Edit Distance}} & \multirow{3}{*}{\parbox{1.1cm}{\centering Total Success Rate}} & \multirow{3}{*}{\parbox{1.1cm}{\centering Unique Success Rate*}} & \multirow{3}{*}{\parbox{1.1cm}{\centering Edit Distance}}  & \multirow{3}{*}{\parbox{1.1cm}{\centering Total Success Rate}} & \multirow{3}{*}{\parbox{1.1cm}{\centering Unique Success Rate*}} & \multirow{3}{*}{\parbox{1.1cm}{\centering Edit Distance}} \\
& & & & & & & & & \\
& & & & & & \\
\midrule
 & \multicolumn{3}{c}{Random} & \multicolumn{3}{|c}{Recombine} & \multicolumn{3}{|c}{Unique Recombine} \\
\midrule
baseline & $8.3$ & $8.2$ & $4.4\pm2.2$ & $36.5$ & $30.0$ & $3.9\pm1.6$ & $39.5$ & $39.5$ & $3.8\pm1.6$ \\
\midrule
 & \multicolumn{3}{c}{SFT} & \multicolumn{3}{|c}{SFT + wBC} & \multicolumn{3}{|c}{SFT + wBC $\backslash$w entropy} \\
random walk 3000 $\times$ 1-hop & $41.3$ & $38.6$ & $4.2\pm2.0$ & $43.6$ & $40.2$ & $4.0\pm1.9$ & $44.3$ & $41.1$\textsuperscript{\ding{61}}  & $4.4\pm2.3$ \\
random walk 1000 $\times$ 3-hop & $41.3$ & $38.6$ & $4.2\pm2.1$ & $43.6$ & $40.1$ & $4.0\pm1.9$ & $44.5$ & $41.2$\textsuperscript{\ding{61}}  & $4.5\pm2.3$ \\
random walk 300 $\times$ 10-hop & $41.8$ & $39.0$ & $4.2\pm2.1$ & $43.8$ & $40.4$ & $4.0\pm1.9$ & $44.7$ & $\mathbf{41.5}$\textsuperscript{\ding{61}}  & $4.5\pm2.4$ \\
priority walk 1000 $\times$ 3-hop & $41.6$ & $38.9$ & $4.2\pm2.0$ & $43.5$ & $40.1$ & $4.0\pm1.9$ & $44.4$ & $41.1$\textsuperscript{\ding{61}}  & $4.5\pm2.3$ \\
priority walk 300 $\times$ 10-hop & $41.1$ & $38.4$ & $4.2\pm2.1$ & $43.5$ & $40.1$ & $4.1\pm1.9$ & $44.6$ & $\mathbf{41.5}$\textsuperscript{\ding{61}} & $4.4\pm2.3$ \\
\bottomrule
\end{tabular}
}
\end{table}

\paragraph{Baselines} We compare our method with two baselines: (1) Random - proteins are randomly mutated based on the edit-distance distribution of the train-set sequences and (2) Recombine - a previous method that samples new diverse sequences by shuffling and merging pairs from an initial seed set \citep{e22090967, sinai2020adalead}. Recombine can sample many duplicate sequences when the seed set is small (certain threshold combinations have fewer than 100 original sequences). We also compare with a stronger variant of Recombine where we ensure that every 3000 newly sampled sequences for a specific multi-attribute constraint are unique and outside the seed set. We call this stronger baseline Unique Recombine. 

\paragraph{{\framework} training} We finetune the ProtGPT2 LM \citep{ferruz2022protgpt2}, which is a 738M parameter protein language model,\footnote{\url{https://huggingface.co/nferruz/ProtGPT2}}, as the rewriter for this task.
Since ProtGPT2 does not have English words as tokens, we prompt the mutant edit pair to the LM as follows: $y_a$\texttt{[A Fluorescence]}$c_1(y_a)$\texttt{[A ddG]}$c_2(y_a)$\texttt{[Target Fluorescence]}$t_{1, a\rightarrow b}$\texttt{[Target ddG]}$t_{2, a\rightarrow b}$\texttt{[edit]}$y_b$, where intermediate key-words are special tokens added to the model's vocabulary. Using the edit pair sampling strategy described earlier, we train ProtGPT2 with SFT for 20K steps with batch size 16 and learning rate $10^{-4}$. We then further continue finetuning with the wBC objective for an additional 10K steps and learning rate $10^{-5}$. Since we want to encourage the LM editor to generate diverse candidates in this task, we separately also train with wBC objective augmented with entropy penalty (coefficient $\gamma = 0.05$).\footnote{$\mathcal{L}_{wBC~\backslash w~entropy}(M) = \mathcal{L}_{wBC}(M) + \gamma (P_M(y_b|x, y_a, C(y_a), T_{a\rightarrow b})\ln P_M(y_b|x, y_a, C(y_a), T_{a\rightarrow b}))$} We present additional implementation details for our methods and the baselines in Appendix \ref{subsec:protein_editors_impl}

\begin{figure}[t]
\begin{subfigure}[t]{0.31\textwidth}
  \centering
  \includegraphics[width=\linewidth]{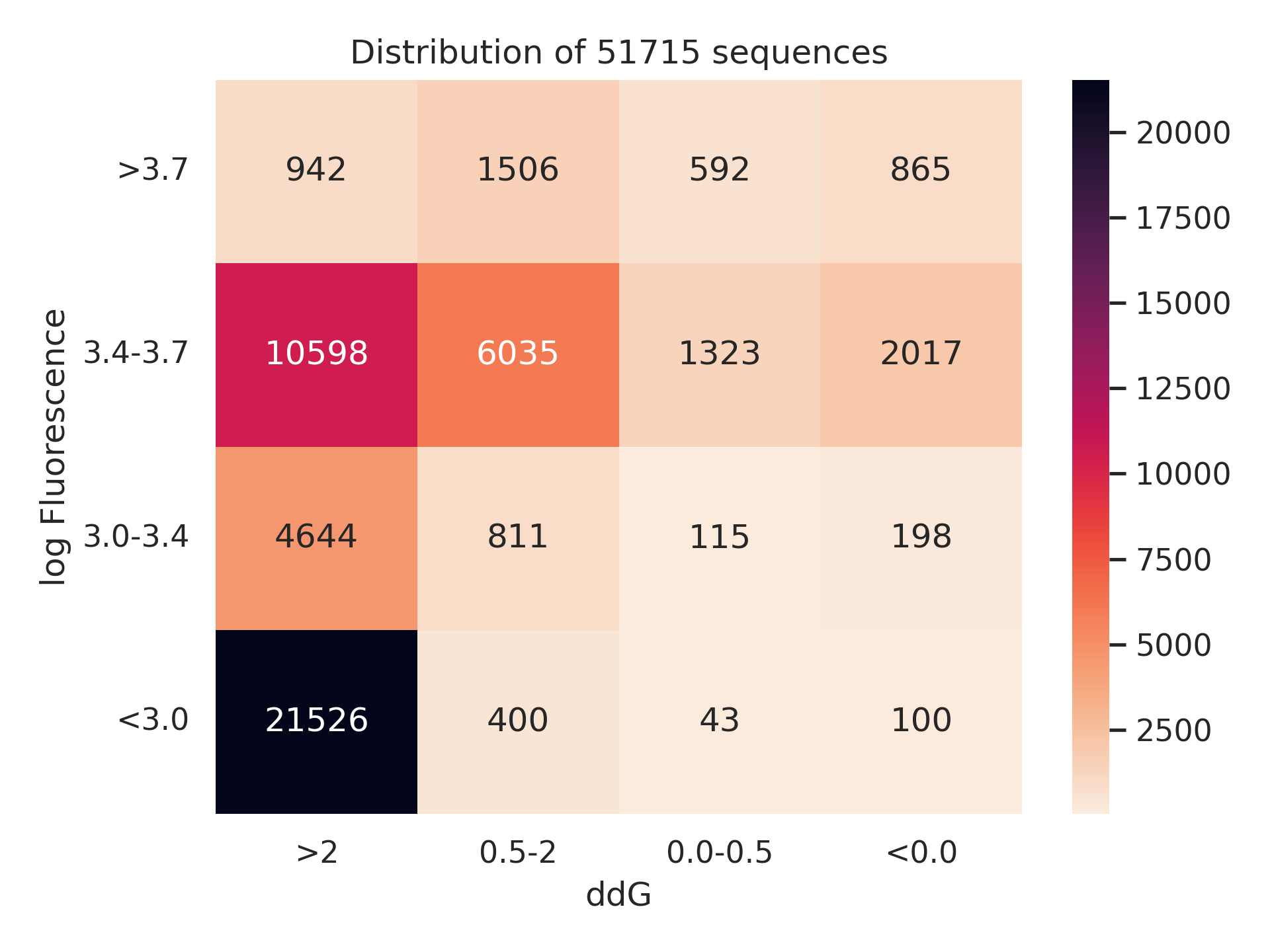}
  \caption{Train Mutant Distribution}
  \label{fig:mut_dist}
\end{subfigure}\hfill
\begin{subfigure}[t]{0.31\textwidth}
  \centering
  \includegraphics[width=\linewidth]{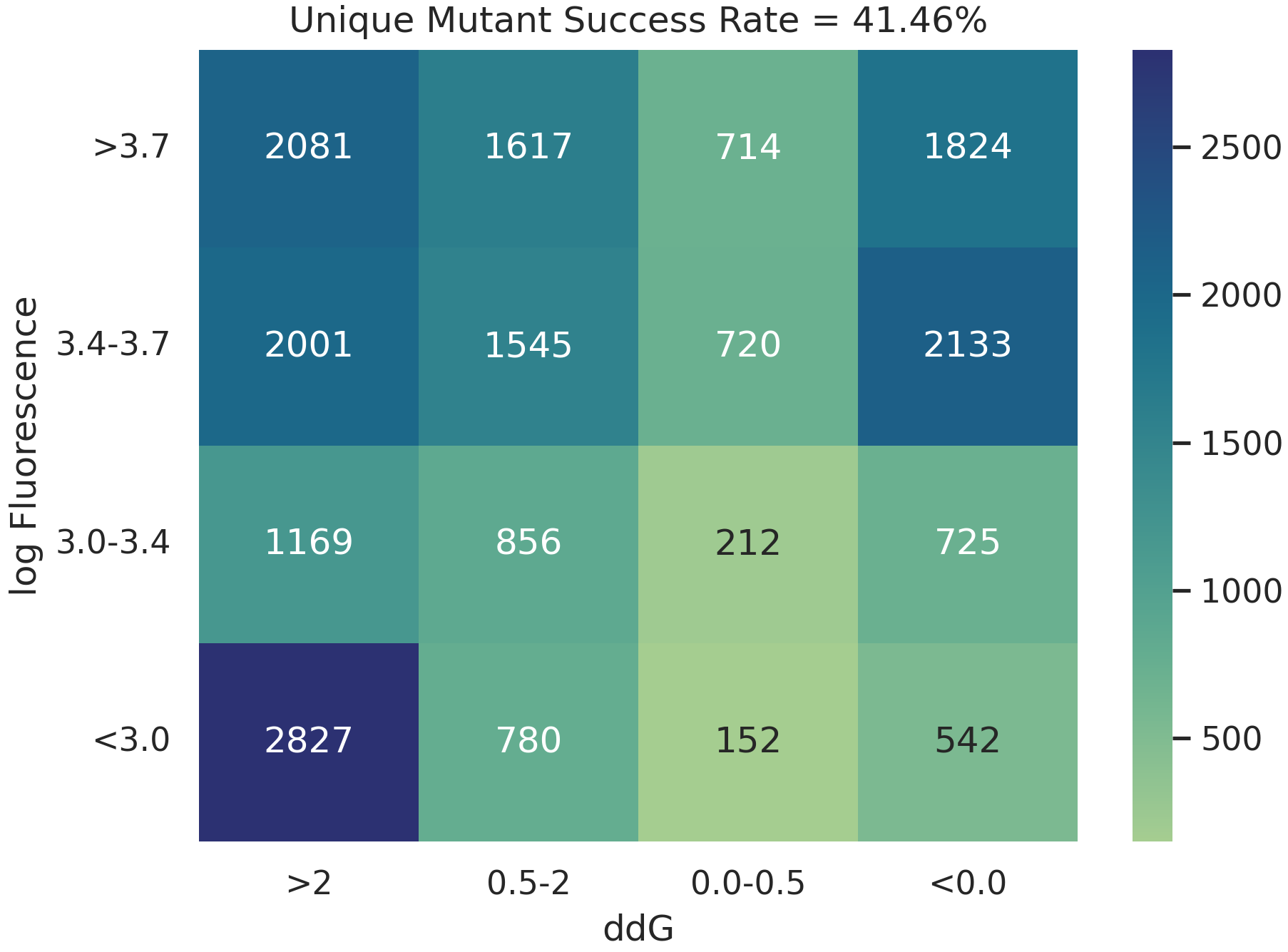}
  \caption{SFT + wBC $\backslash$w entropy - priority 300 $\times$ 10-hop} 
  \label{fig:protein_nll_wbc_entorpy_best_of_N}
\end{subfigure}
\begin{subfigure}[t]{0.37\textwidth}
  \centering
  \includegraphics[width=\linewidth]{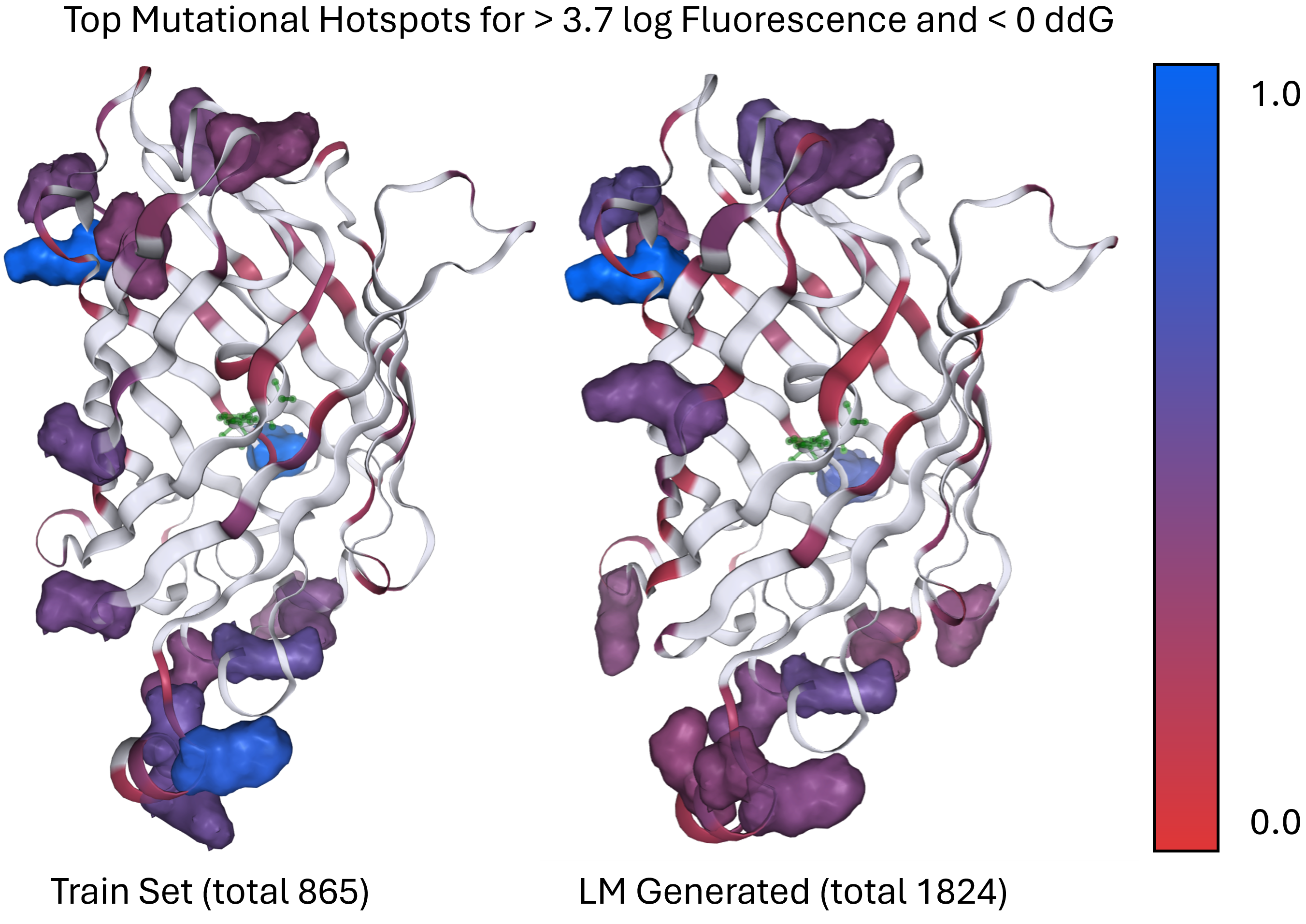}
  \caption{Comparing Mutational Hotspots} 
  \label{fig:protien_mut_hotspots}
\end{subfigure}
\caption{Analyzing new mutant discovery of reward-prioritized walk compared with training set distribution. \textbf{\textit{Takeaway}:} Even with very few training instances in many of the regions, LMs trained with {\framework} discover many novel candidates. Analysis of the top 15 mutational hotspots reveals that LMs can extrapolate beyond the mutational patterns seen in the training set.}
\label{fig:unique_mutants_analysis}
\end{figure}

\subsection{Protein Design Results}
We report the evaluation results of baselines and different variants of {\framework} in Table \ref{tab:pd_main_result}. Random mutation shows the worst performance as expected, while Recombine is a strong baseline that finds more unique and successful mutants. With Unique Recombine, we establish an upper bound on the baseline's performance by only retaining unique sequences. However, the offline wBC model outperforms both the Recombine baseline and the SFT model across all inference strategies. When augmented with entropy penalty, we observe a boost in success rate for the wBC model and a larger spread of edit distances indicating more diverse mutants. Although reward-prioritized and naive multi-hop walks yield the best performance, we do not notice a significant difference between different inference strategies with two proportions z-test. 

Finally, we compare the distribution of train set mutant vs. the newly discovered mutants via the reward-prioritized walk (300 $\times$ 10-hop) from wBC $\backslash$w entropy model in Figure \ref{fig:unique_mutants_analysis}. Despite small sample sizes of training data in certain regions, our method can extrapolate beyond the original training set and find diverse sequences even with offline training. Finally, when comparing mutational hotspots and their distribution across GFP structure, our LM-generated sequences show a different distribution and occasionally novel mutations compared to train set sequences, as shown in Figure \ref{fig:protien_mut_hotspots} (and Figure \ref{fig:extra_protein_hotspots} in the Appendix).

\section{Conclusion} 
We create {\frameworkLong} (\framework) framework to cheaply train LMs as fine-grained editors by sampling edit pairs from offline sequential datasets. We also create a new {\benchmarkLong} ({\benchmark}) benchmark to evaluate our method, comprising two challenging fine-grained controllability tasks. In the {\benchmark} Text Style Transfer task, LM editors trained with weighted behavior cloning paired with proposed k-NN edit pair sampling, and multi-step reward-prioritized editing outperform their SFT counterparts and other inference methods. We boost its performance further with anchor conditioning and achieve the highest constraint satisfaction rates compared to previous fine-tuning and few-shot prompted baselines. Interestingly, in the {\benchmark} Protein Design task, {\framework} can train protein language models to discover novel proteins outside the training data with high success rates while highlighting different mutational hotspots. Our study demonstrates the potential of LMs as fine-grained writing assistants and protein engineering models that can aid in the creation of novel proteins with fine-grained properties.

\bibliography{anthology_filtered,custom}

\begin{thebibliography}{85}
\providecommand{\natexlab}[1]{#1}
\providecommand{\url}[1]{\texttt{#1}}
\expandafter\ifx\csname urlstyle\endcsname\relax
  \providecommand{\doi}[1]{doi: #1}\else
  \providecommand{\doi}{doi: \begingroup \urlstyle{rm}\Url}\fi

\bibitem[AI@Meta(2024)]{llama3modelcard}
AI@Meta.
\newblock Llama 3 model card.
\newblock 2024.
\newblock URL \url{https://github.com/meta-llama/llama3/blob/main/MODEL_CARD.md}.

\bibitem[Akyurek et~al.(2023)Akyurek, Akyurek, Kalyan, Clark, Wijaya, and Tandon]{akyurek-etal-2023-rl4f}
Afra~Feyza Akyurek, Ekin Akyurek, Ashwin Kalyan, Peter Clark, Derry~Tanti Wijaya, and Niket Tandon.
\newblock {RL}4{F}: Generating natural language feedback with reinforcement learning for repairing model outputs.
\newblock In Anna Rogers, Jordan Boyd-Graber, and Naoaki Okazaki (eds.), \emph{Proceedings of the 61st Annual Meeting of the Association for Computational Linguistics (Volume 1: Long Papers)}, pp.\  7716--7733, Toronto, Canada, July 2023. Association for Computational Linguistics.
\newblock \doi{10.18653/v1/2023.acl-long.427}.
\newblock URL \url{https://aclanthology.org/2023.acl-long.427}.

\bibitem[Amat \& Nifosì(2013)Amat and Nifosì]{10.1021_ct3007452}
Pietro Amat and Riccardo Nifosì.
\newblock Spectral “fine” tuning in fluorescent proteins: The case of the gfp-like chromophore in the anionic protonation state.
\newblock \emph{Journal of Chemical Theory and Computation}, 9\penalty0 (1):\penalty0 497--508, 2013.
\newblock \doi{10.1021/ct3007452}.
\newblock URL \url{https://doi.org/10.1021/ct3007452}.
\newblock PMID: 26589050.

\bibitem[Andrychowicz et~al.(2017)Andrychowicz, Wolski, Ray, Schneider, Fong, Welinder, McGrew, Tobin, Pieter~Abbeel, and Zaremba]{NIPS2017_453fadbd}
Marcin Andrychowicz, Filip Wolski, Alex Ray, Jonas Schneider, Rachel Fong, Peter Welinder, Bob McGrew, Josh Tobin, OpenAI Pieter~Abbeel, and Wojciech Zaremba.
\newblock Hindsight experience replay.
\newblock In I.~Guyon, U.~Von Luxburg, S.~Bengio, H.~Wallach, R.~Fergus, S.~Vishwanathan, and R.~Garnett (eds.), \emph{Advances in Neural Information Processing Systems}, volume~30. Curran Associates, Inc., 2017.
\newblock URL \url{https://proceedings.neurips.cc/paper_files/paper/2017/file/453fadbd8a1a3af50a9df4df899537b5-Paper.pdf}.

\bibitem[Baheti et~al.(2024)Baheti, Lu, Brahman, Bras, Sap, and Riedl]{baheti2024leftover}
Ashutosh Baheti, Ximing Lu, Faeze Brahman, Ronan~Le Bras, Maarten Sap, and Mark Riedl.
\newblock Leftover lunch: Advantage-based offline reinforcement learning for language models.
\newblock In \emph{The Twelfth International Conference on Learning Representations}, 2024.
\newblock URL \url{https://openreview.net/forum?id=ZDGKPbF0VQ}.

\bibitem[Bolognesi et~al.(2019)Bolognesi, Faure, Seuma, Schmiedel, Tartaglia, and Lehner]{bolognesi2019mutational}
Benedetta Bolognesi, Andre~J Faure, Mireia Seuma, J{\"o}rn~M Schmiedel, Gian~Gaetano Tartaglia, and Ben Lehner.
\newblock The mutational landscape of a prion-like domain.
\newblock \emph{Nature communications}, 10\penalty0 (1):\penalty0 4162, 2019.

\bibitem[Bryant et~al.(2021)Bryant, Bashir, Sinai, Jain, Ogden, Riley, Church, Colwell, and Kelsic]{bryant2021deep}
Drew~H Bryant, Ali Bashir, Sam Sinai, Nina~K Jain, Pierce~J Ogden, Patrick~F Riley, George~M Church, Lucy~J Colwell, and Eric~D Kelsic.
\newblock Deep diversification of an aav capsid protein by machine learning.
\newblock \emph{Nature Biotechnology}, 39\penalty0 (6):\penalty0 691--696, 2021.

\bibitem[Chan et~al.(2021)Chan, Madani, Krause, and Naik]{chan2021deep}
Alvin Chan, Ali Madani, Ben Krause, and Nikhil Naik.
\newblock Deep extrapolation for attribute-enhanced generation.
\newblock In A.~Beygelzimer, Y.~Dauphin, P.~Liang, and J.~Wortman Vaughan (eds.), \emph{Advances in Neural Information Processing Systems}, 2021.
\newblock URL \url{https://openreview.net/forum?id=NCDMYD2y5kK}.

\bibitem[Childers \& Daggett(2017)Childers and Daggett]{childers2017insights}
Matthew~Carter Childers and Valerie Daggett.
\newblock Insights from molecular dynamics simulations for computational protein design.
\newblock \emph{Molecular systems design \& engineering}, 2\penalty0 (1):\penalty0 9--33, 2017.

\bibitem[Collins \& Koo(2005)Collins and Koo]{10.1162/0891201053630273}
Michael Collins and Terry Koo.
\newblock {Discriminative Reranking for Natural Language Parsing}.
\newblock \emph{Computational Linguistics}, 31\penalty0 (1):\penalty0 25--70, 03 2005.
\newblock ISSN 0891-2017.
\newblock \doi{10.1162/0891201053630273}.
\newblock URL \url{https://doi.org/10.1162/0891201053630273}.

\bibitem[Dallago et~al.(2021)Dallago, Mou, Johnston, Wittmann, Bhattacharya, Goldman, Madani, and Yang]{dallago2021flip}
Christian Dallago, Jody Mou, Kadina~E Johnston, Bruce Wittmann, Nick Bhattacharya, Samuel Goldman, Ali Madani, and Kevin~K Yang.
\newblock {FLIP}: Benchmark tasks in fitness landscape inference for proteins.
\newblock In \emph{Thirty-fifth Conference on Neural Information Processing Systems Datasets and Benchmarks Track (Round 2)}, 2021.
\newblock URL \url{https://openreview.net/forum?id=p2dMLEwL8tF}.

\bibitem[Dathathri et~al.(2020)Dathathri, Madotto, Lan, Hung, Frank, Molino, Yosinski, and Liu]{Dathathri2020Plug}
Sumanth Dathathri, Andrea Madotto, Janice Lan, Jane Hung, Eric Frank, Piero Molino, Jason Yosinski, and Rosanne Liu.
\newblock Plug and play language models: A simple approach to controlled text generation.
\newblock In \emph{International Conference on Learning Representations}, 2020.
\newblock URL \url{https://openreview.net/forum?id=H1edEyBKDS}.

\bibitem[Dhuliawala et~al.(2023)Dhuliawala, Komeili, Xu, Raileanu, Li, Celikyilmaz, and Weston]{dhuliawala2023chainofverification}
Shehzaad Dhuliawala, Mojtaba Komeili, Jing Xu, Roberta Raileanu, Xian Li, Asli Celikyilmaz, and Jason Weston.
\newblock Chain-of-verification reduces hallucination in large language models, 2023.

\bibitem[Dill et~al.(2008)Dill, Ozkan, Shell, and Weikl]{doi:10.1146/annurev.biophys.37.092707.153558}
Ken~A. Dill, S.~Banu Ozkan, M.~Scott Shell, and Thomas~R. Weikl.
\newblock The protein folding problem.
\newblock \emph{Annual Review of Biophysics}, 37\penalty0 (1):\penalty0 289--316, 2008.
\newblock \doi{10.1146/annurev.biophys.37.092707.153558}.
\newblock URL \url{https://doi.org/10.1146/annurev.biophys.37.092707.153558}.
\newblock PMID: 18573083.

\bibitem[Ding et~al.(2023)Ding, Pang, Wei, Shen, Cheng, and Chua]{ding2023maclasa}
Hanxing Ding, Liang Pang, Zihao Wei, Huawei Shen, Xueqi Cheng, and Tat-Seng Chua.
\newblock Maclasa: Multi-aspect controllable text generation via efficient sampling from compact latent space, 2023.

\bibitem[Feng et~al.(2023)Feng, Yang, Zhang, Zhang, Xiong, Zhou, and Wang]{feng2023fantastic}
Yihao Feng, Shentao Yang, Shujian Zhang, Jianguo Zhang, Caiming Xiong, Mingyuan Zhou, and Huan Wang.
\newblock Fantastic rewards and how to tame them: A case study on reward learning for task-oriented dialogue systems.
\newblock In \emph{The Eleventh International Conference on Learning Representations}, 2023.
\newblock URL \url{https://openreview.net/forum?id=086pmarAris}.

\bibitem[Ferruz et~al.(2022)Ferruz, Schmidt, and H{\"o}cker]{ferruz2022protgpt2}
Noelia Ferruz, Steffen Schmidt, and Birte H{\"o}cker.
\newblock Protgpt2 is a deep unsupervised language model for protein design.
\newblock \emph{Nature communications}, 13\penalty0 (1):\penalty0 4348, 2022.

\bibitem[Fleiss et~al.(2013)Fleiss, Levin, and Paik]{fleiss2013statistical}
Joseph~L Fleiss, Bruce Levin, and Myunghee~Cho Paik.
\newblock \emph{Statistical methods for rates and proportions}.
\newblock john wiley \& sons, 2013.

\bibitem[Garcia~Seisdedos et~al.(2022)Garcia~Seisdedos, Levin, Shapira, Freud, and Levy]{garcia2022mutant}
Hector Garcia~Seisdedos, Tal Levin, Gal Shapira, Saskia Freud, and Emmanuel~D Levy.
\newblock Mutant libraries reveal negative design shielding proteins from supramolecular self-assembly and relocalization in cells.
\newblock \emph{Proceedings of the National Academy of Sciences}, 119\penalty0 (5):\penalty0 e2101117119, 2022.

\bibitem[Ghosh et~al.(2021)Ghosh, Qi, Chaturvedi, and Srivastava]{ghosh-etal-2021-helpful}
Sayan Ghosh, Zheng Qi, Snigdha Chaturvedi, and Shashank Srivastava.
\newblock How helpful is inverse reinforcement learning for table-to-text generation?
\newblock In Chengqing Zong, Fei Xia, Wenjie Li, and Roberto Navigli (eds.), \emph{Proceedings of the 59th Annual Meeting of the Association for Computational Linguistics and the 11th International Joint Conference on Natural Language Processing (Volume 2: Short Papers)}, pp.\  71--79, Online, August 2021. Association for Computational Linguistics.
\newblock \doi{10.18653/v1/2021.acl-short.11}.
\newblock URL \url{https://aclanthology.org/2021.acl-short.11}.

\bibitem[Gonzalez~Somermeyer et~al.(2022)Gonzalez~Somermeyer, Fleiss, Mishin, Bozhanova, Igolkina, Meiler, Alaball~Pujol, Putintseva, Sarkisyan, and Kondrashov]{10.7554/eLife.75842}
Louisa Gonzalez~Somermeyer, Aubin Fleiss, Alexander~S Mishin, Nina~G Bozhanova, Anna~A Igolkina, Jens Meiler, Maria-Elisenda Alaball~Pujol, Ekaterina~V Putintseva, Karen~S Sarkisyan, and Fyodor~A Kondrashov.
\newblock Heterogeneity of the gfp fitness landscape and data-driven protein design.
\newblock \emph{eLife}, 11:\penalty0 e75842, may 2022.
\newblock ISSN 2050-084X.
\newblock \doi{10.7554/eLife.75842}.
\newblock URL \url{https://doi.org/10.7554/eLife.75842}.

\bibitem[Gou et~al.(2024)Gou, Shao, Gong, yelong shen, Yang, Duan, and Chen]{gou2024critic}
Zhibin Gou, Zhihong Shao, Yeyun Gong, yelong shen, Yujiu Yang, Nan Duan, and Weizhu Chen.
\newblock {CRITIC}: Large language models can self-correct with tool-interactive critiquing.
\newblock In \emph{The Twelfth International Conference on Learning Representations}, 2024.
\newblock URL \url{https://openreview.net/forum?id=Sx038qxjek}.

\bibitem[Gu et~al.(2022)Gu, Feng, Ma, Zhang, Gong, and Qin]{gu-etal-2022-distributional}
Yuxuan Gu, Xiaocheng Feng, Sicheng Ma, Lingyuan Zhang, Heng Gong, and Bing Qin.
\newblock A distributional lens for multi-aspect controllable text generation.
\newblock In Yoav Goldberg, Zornitsa Kozareva, and Yue Zhang (eds.), \emph{Proceedings of the 2022 Conference on Empirical Methods in Natural Language Processing}, pp.\  1023--1043, Abu Dhabi, United Arab Emirates, December 2022. Association for Computational Linguistics.
\newblock \doi{10.18653/v1/2022.emnlp-main.67}.
\newblock URL \url{https://aclanthology.org/2022.emnlp-main.67}.

\bibitem[Guo et~al.(2004)Guo, Choe, and Loeb]{Guo2004ProteinTT}
Haiwei~Henry Guo, Juno Choe, and Lawrence~A. Loeb.
\newblock Protein tolerance to random amino acid change.
\newblock \emph{Proceedings of the National Academy of Sciences of the United States of America}, 101 25:\penalty0 9205--10, 2004.
\newblock URL \url{https://api.semanticscholar.org/CorpusID:7391571}.

\bibitem[Hallinan et~al.(2023)Hallinan, Brahman, Lu, Jung, Welleck, and Choi]{hallinan-etal-2023-steer}
Skyler Hallinan, Faeze Brahman, Ximing Lu, Jaehun Jung, Sean Welleck, and Yejin Choi.
\newblock {STEER}: Unified style transfer with expert reinforcement.
\newblock In Houda Bouamor, Juan Pino, and Kalika Bali (eds.), \emph{Findings of the Association for Computational Linguistics: EMNLP 2023}, pp.\  7546--7562, Singapore, December 2023. Association for Computational Linguistics.
\newblock \doi{10.18653/v1/2023.findings-emnlp.506}.
\newblock URL \url{https://aclanthology.org/2023.findings-emnlp.506}.

\bibitem[He et~al.(2020)He, Wang, Neubig, and Berg-Kirkpatrick]{He2020A}
Junxian He, Xinyi Wang, Graham Neubig, and Taylor Berg-Kirkpatrick.
\newblock A probabilistic formulation of unsupervised text style transfer.
\newblock In \emph{International Conference on Learning Representations}, 2020.
\newblock URL \url{https://openreview.net/forum?id=HJlA0C4tPS}.

\bibitem[Holtzman et~al.(2019)Holtzman, Buys, Du, Forbes, and Choi]{holtzman2019curious}
Ari Holtzman, Jan Buys, Li~Du, Maxwell Forbes, and Yejin Choi.
\newblock The curious case of neural text degeneration.
\newblock In \emph{International Conference on Learning Representations}, 2019.

\bibitem[Hsu et~al.(2022)Hsu, Nisonoff, Fannjiang, and Listgarten]{Hsu2022}
Chloe Hsu, Hunter Nisonoff, Clara Fannjiang, and Jennifer Listgarten.
\newblock Learning protein fitness models from evolutionary and assay-labeled data.
\newblock \emph{Nature Biotechnology}, 40\penalty0 (7):\penalty0 1114--1122, Jul 2022.
\newblock ISSN 1546-1696.
\newblock \doi{10.1038/s41587-021-01146-5}.
\newblock URL \url{https://doi.org/10.1038/s41587-021-01146-5}.

\bibitem[Hu et~al.(2023)Hu, Cao, Chan, Liu, Xiao, Su, and Wu]{9944920}
Zhe Hu, Zhiwei Cao, Hou~Pong Chan, Jiachen Liu, Xinyan Xiao, Jinsong Su, and Hua Wu.
\newblock Controllable dialogue generation with disentangled multi-grained style specification and attribute consistency reward.
\newblock \emph{IEEE/ACM Transactions on Audio, Speech, and Language Processing}, 31:\penalty0 188--199, 2023.
\newblock \doi{10.1109/TASLP.2022.3221002}.

\bibitem[Huang et~al.(1996)Huang, Petrosino, Hirsch, Shenkin, and Palzkill]{HUANG1996688}
Wanzhi Huang, Joseph Petrosino, Marc Hirsch, Peter~S. Shenkin, and Timothy Palzkill.
\newblock Amino acid sequence determinants of $\beta$-lactamase structure and activity.
\newblock \emph{Journal of Molecular Biology}, 258\penalty0 (4):\penalty0 688--703, 1996.
\newblock ISSN 0022-2836.
\newblock \doi{https://doi.org/10.1006/jmbi.1996.0279}.
\newblock URL \url{https://www.sciencedirect.com/science/article/pii/S002228369690279X}.

\bibitem[Jain et~al.(2022)Jain, Bengio, Hernandez-Garcia, Rector-Brooks, Dossou, Ekbote, Fu, Zhang, Kilgour, Zhang, Simine, Das, and Bengio]{pmlr-v162-jain22a}
Moksh Jain, Emmanuel Bengio, Alex Hernandez-Garcia, Jarrid Rector-Brooks, Bonaventure F.~P. Dossou, Chanakya~Ajit Ekbote, Jie Fu, Tianyu Zhang, Michael Kilgour, Dinghuai Zhang, Lena Simine, Payel Das, and Yoshua Bengio.
\newblock Biological sequence design with {GF}low{N}ets.
\newblock In Kamalika Chaudhuri, Stefanie Jegelka, Le~Song, Csaba Szepesvari, Gang Niu, and Sivan Sabato (eds.), \emph{Proceedings of the 39th International Conference on Machine Learning}, volume 162 of \emph{Proceedings of Machine Learning Research}, pp.\  9786--9801. PMLR, 17--23 Jul 2022.
\newblock URL \url{https://proceedings.mlr.press/v162/jain22a.html}.

\bibitem[Junczys-Dowmunt et~al.(2018)Junczys-Dowmunt, Grundkiewicz, Guha, and Heafield]{junczys-dowmunt-etal-2018-approaching}
Marcin Junczys-Dowmunt, Roman Grundkiewicz, Shubha Guha, and Kenneth Heafield.
\newblock Approaching neural grammatical error correction as a low-resource machine translation task.
\newblock In Marilyn Walker, Heng Ji, and Amanda Stent (eds.), \emph{Proceedings of the 2018 Conference of the North {A}merican Chapter of the Association for Computational Linguistics: Human Language Technologies, Volume 1 (Long Papers)}, pp.\  595--606, New Orleans, Louisiana, June 2018. Association for Computational Linguistics.
\newblock \doi{10.18653/v1/N18-1055}.
\newblock URL \url{https://aclanthology.org/N18-1055}.

\bibitem[Keskar et~al.(2019)Keskar, McCann, Varshney, Xiong, and Socher]{keskarCTRL2019}
Nitish~Shirish Keskar, Bryan McCann, Lav Varshney, Caiming Xiong, and Richard Socher.
\newblock {CTRL - A Conditional Transformer Language Model for Controllable Generation}.
\newblock \emph{arXiv preprint arXiv:1909.05858}, 2019.

\bibitem[Kirjner et~al.(2024)Kirjner, Yim, Samusevich, Bracha, Jaakkola, Barzilay, and Fiete]{kirjner2024improving}
Andrew Kirjner, Jason Yim, Raman Samusevich, Shahar Bracha, Tommi~S. Jaakkola, Regina Barzilay, and Ila~R Fiete.
\newblock Improving protein optimization with smoothed fitness landscapes.
\newblock In \emph{The Twelfth International Conference on Learning Representations}, 2024.
\newblock URL \url{https://openreview.net/forum?id=rxlF2Zv8x0}.

\bibitem[Krishna et~al.(2020)Krishna, Wieting, and Iyyer]{krishna-etal-2020-reformulating}
Kalpesh Krishna, John Wieting, and Mohit Iyyer.
\newblock Reformulating unsupervised style transfer as paraphrase generation.
\newblock In Bonnie Webber, Trevor Cohn, Yulan He, and Yang Liu (eds.), \emph{Proceedings of the 2020 Conference on Empirical Methods in Natural Language Processing (EMNLP)}, pp.\  737--762, Online, November 2020. Association for Computational Linguistics.
\newblock \doi{10.18653/v1/2020.emnlp-main.55}.
\newblock URL \url{https://aclanthology.org/2020.emnlp-main.55}.

\bibitem[Kumar et~al.(2021)Kumar, Malmi, Severyn, and Tsvetkov]{NEURIPS2021_79ec2a42}
Sachin Kumar, Eric Malmi, Aliaksei Severyn, and Yulia Tsvetkov.
\newblock Controlled text generation as continuous optimization with multiple constraints.
\newblock In M.~Ranzato, A.~Beygelzimer, Y.~Dauphin, P.S. Liang, and J.~Wortman Vaughan (eds.), \emph{Advances in Neural Information Processing Systems}, volume~34, pp.\  14542--14554. Curran Associates, Inc., 2021.
\newblock URL \url{https://proceedings.neurips.cc/paper_files/paper/2021/file/79ec2a4246feb2126ecf43c4a4418002-Paper.pdf}.

\bibitem[Kumar et~al.(2022)Kumar, Paria, and Tsvetkov]{kumar-etal-2022-gradient}
Sachin Kumar, Biswajit Paria, and Yulia Tsvetkov.
\newblock Gradient-based constrained sampling from language models.
\newblock In Yoav Goldberg, Zornitsa Kozareva, and Yue Zhang (eds.), \emph{Proceedings of the 2022 Conference on Empirical Methods in Natural Language Processing}, pp.\  2251--2277, Abu Dhabi, United Arab Emirates, December 2022. Association for Computational Linguistics.
\newblock \doi{10.18653/v1/2022.emnlp-main.144}.
\newblock URL \url{https://aclanthology.org/2022.emnlp-main.144}.

\bibitem[Laban et~al.(2023)Laban, Vig, Kryscinski, Joty, Xiong, and Wu]{laban-etal-2023-swipe}
Philippe Laban, Jesse Vig, Wojciech Kryscinski, Shafiq Joty, Caiming Xiong, and Chien-Sheng Wu.
\newblock {SW}i{PE}: A dataset for document-level simplification of {W}ikipedia pages.
\newblock In Anna Rogers, Jordan Boyd-Graber, and Naoaki Okazaki (eds.), \emph{Proceedings of the 61st Annual Meeting of the Association for Computational Linguistics (Volume 1: Long Papers)}, pp.\  10674--10695, Toronto, Canada, July 2023. Association for Computational Linguistics.
\newblock \doi{10.18653/v1/2023.acl-long.596}.
\newblock URL \url{https://aclanthology.org/2023.acl-long.596}.

\bibitem[Lample et~al.(2019)Lample, Subramanian, Smith, Denoyer, Ranzato, and Boureau]{lample2018multipleattribute}
Guillaume Lample, Sandeep Subramanian, Eric Smith, Ludovic Denoyer, Marc'Aurelio Ranzato, and Y-Lan Boureau.
\newblock Multiple-attribute text rewriting.
\newblock In \emph{International Conference on Learning Representations}, 2019.
\newblock URL \url{https://openreview.net/forum?id=H1g2NhC5KQ}.

\bibitem[Li et~al.(2018)Li, Jia, He, and Liang]{li-etal-2018-delete}
Juncen Li, Robin Jia, He~He, and Percy Liang.
\newblock Delete, retrieve, generate: a simple approach to sentiment and style transfer.
\newblock In Marilyn Walker, Heng Ji, and Amanda Stent (eds.), \emph{Proceedings of the 2018 Conference of the North {A}merican Chapter of the Association for Computational Linguistics: Human Language Technologies, Volume 1 (Long Papers)}, pp.\  1865--1874, New Orleans, Louisiana, June 2018. Association for Computational Linguistics.
\newblock \doi{10.18653/v1/N18-1169}.
\newblock URL \url{https://aclanthology.org/N18-1169}.

\bibitem[Li et~al.(2022)Li, Thickstun, Gulrajani, Liang, and Hashimoto]{NEURIPS2022_1be5bc25}
Xiang Li, John Thickstun, Ishaan Gulrajani, Percy~S Liang, and Tatsunori~B Hashimoto.
\newblock Diffusion-lm improves controllable text generation.
\newblock In S.~Koyejo, S.~Mohamed, A.~Agarwal, D.~Belgrave, K.~Cho, and A.~Oh (eds.), \emph{Advances in Neural Information Processing Systems}, volume~35, pp.\  4328--4343. Curran Associates, Inc., 2022.
\newblock URL \url{https://proceedings.neurips.cc/paper_files/paper/2022/file/1be5bc25d50895ee656b8c2d9eb89d6a-Paper-Conference.pdf}.

\bibitem[Lin et~al.(2023)Lin, Akin, Rao, Hie, Zhu, Lu, Smetanin, Verkuil, Kabeli, Shmueli, dos Santos~Costa, Fazel-Zarandi, Sercu, Candido, and Rives]{lin2023evolutionary}
Zeming Lin, Halil Akin, Roshan Rao, Brian Hie, Zhongkai Zhu, Wenting Lu, Nikita Smetanin, Robert Verkuil, Ori Kabeli, Yaniv Shmueli, Allan dos Santos~Costa, Maryam Fazel-Zarandi, Tom Sercu, Salvatore Candido, and Alexander Rives.
\newblock Evolutionary-scale prediction of atomic-level protein structure with a language model.
\newblock \emph{Science}, 379\penalty0 (6637):\penalty0 1123--1130, 2023.
\newblock \doi{10.1126/science.ade2574}.
\newblock URL \url{https://www.science.org/doi/abs/10.1126/science.ade2574}.
\newblock Earlier versions as preprint: bioRxiv 2022.07.20.500902.

\bibitem[Liu et~al.(2022)Liu, Feng, Gao, Yang, Liang, Bao, He, Cui, Li, and Hu]{liu2022composable}
Guangyi Liu, Zeyu Feng, Yuan Gao, Zichao Yang, Xiaodan Liang, Junwei Bao, Xiaodong He, Shuguang Cui, Zhen Li, and Zhiting Hu.
\newblock Composable text control operations in latent space with ordinary differential equations.
\newblock \emph{arXiv preprint arXiv:2208.00638}, 2022.

\bibitem[Liu et~al.(2023)Liu, Khalifa, and Wang]{liu-etal-2023-bolt}
Xin Liu, Muhammad Khalifa, and Lu~Wang.
\newblock {BOLT}: Fast energy-based controlled text generation with tunable biases.
\newblock In Anna Rogers, Jordan Boyd-Graber, and Naoaki Okazaki (eds.), \emph{Proceedings of the 61st Annual Meeting of the Association for Computational Linguistics (Volume 2: Short Papers)}, pp.\  186--200, Toronto, Canada, July 2023. Association for Computational Linguistics.
\newblock \doi{10.18653/v1/2023.acl-short.18}.
\newblock URL \url{https://aclanthology.org/2023.acl-short.18}.

\bibitem[Liu et~al.(2020)Liu, Ott, Goyal, Du, Joshi, Chen, Levy, Lewis, Zettlemoyer, and Stoyanov]{liu2020roberta}
Yinhan Liu, Myle Ott, Naman Goyal, Jingfei Du, Mandar Joshi, Danqi Chen, Omer Levy, Mike Lewis, Luke Zettlemoyer, and Veselin Stoyanov.
\newblock Ro{\{}bert{\}}a: A robustly optimized {\{}bert{\}} pretraining approach, 2020.
\newblock URL \url{https://openreview.net/forum?id=SyxS0T4tvS}.

\bibitem[Lu et~al.(2022)Lu, Welleck, Hessel, Jiang, Qin, West, Ammanabrolu, and Choi]{lu2022quark}
Ximing Lu, Sean Welleck, Jack Hessel, Liwei Jiang, Lianhui Qin, Peter West, Prithviraj Ammanabrolu, and Yejin Choi.
\newblock {QUARK}: Controllable text generation with reinforced unlearning.
\newblock In Alice~H. Oh, Alekh Agarwal, Danielle Belgrave, and Kyunghyun Cho (eds.), \emph{Advances in Neural Information Processing Systems}, 2022.
\newblock URL \url{https://openreview.net/forum?id=5HaIds3ux5O}.

\bibitem[Luo et~al.(2019)Luo, Li, Zhou, Yang, Chang, Sui, and Sun]{Luo19DualRL}
Fuli Luo, Peng Li, Jie Zhou, Pengcheng Yang, Baobao Chang, Zhifang Sui, and Xu~Sun.
\newblock A dual reinforcement learning framework for unsupervised text style transfer.
\newblock In \emph{Proceedings of the 28th International Joint Conference on Artificial Intelligence, {IJCAI} 2019}, 2019.

\bibitem[Ma et~al.(2020)Ma, Sap, Rashkin, and Choi]{ma2020powerTransformer}
Xinyao Ma, Maarten Sap, Hannah Rashkin, and Yejin Choi.
\newblock Powertransformer: Unsupervised controllable revision for biased language correction.
\newblock In \emph{EMNLP}, 2020.
\newblock URL \url{https://www.aclweb.org/anthology/2020.emnlp-main.602}.

\bibitem[Madaan et~al.(2023)Madaan, Tandon, Gupta, Hallinan, Gao, Wiegreffe, Alon, Dziri, Prabhumoye, Yang, Welleck, Majumder, Gupta, Yazdanbakhsh, and Clark]{madaan2023selfrefine}
Aman Madaan, Niket Tandon, Prakhar Gupta, Skyler Hallinan, Luyu Gao, Sarah Wiegreffe, Uri Alon, Nouha Dziri, Shrimai Prabhumoye, Yiming Yang, Sean Welleck, Bodhisattwa~Prasad Majumder, Shashank Gupta, Amir Yazdanbakhsh, and Peter Clark.
\newblock Self-refine: Iterative refinement with self-feedback, 2023.

\bibitem[Mireshghallah et~al.(2022)Mireshghallah, Goyal, and Berg-Kirkpatrick]{mireshghallah-etal-2022-mix}
Fatemehsadat Mireshghallah, Kartik Goyal, and Taylor Berg-Kirkpatrick.
\newblock Mix and match: Learning-free controllable text generationusing energy language models.
\newblock In Smaranda Muresan, Preslav Nakov, and Aline Villavicencio (eds.), \emph{Proceedings of the 60th Annual Meeting of the Association for Computational Linguistics (Volume 1: Long Papers)}, pp.\  401--415, Dublin, Ireland, May 2022. Association for Computational Linguistics.
\newblock \doi{10.18653/v1/2022.acl-long.31}.
\newblock URL \url{https://aclanthology.org/2022.acl-long.31}.

\bibitem[Norouzi et~al.(2016)Norouzi, Bengio, Chen, Jaitly, Schuster, Wu, and Schuurmans]{NIPS2016_2f885d0f}
Mohammad Norouzi, Samy Bengio, zhifeng Chen, Navdeep Jaitly, Mike Schuster, Yonghui Wu, and Dale Schuurmans.
\newblock Reward augmented maximum likelihood for neural structured prediction.
\newblock In D.~Lee, M.~Sugiyama, U.~Luxburg, I.~Guyon, and R.~Garnett (eds.), \emph{Advances in Neural Information Processing Systems}, volume~29. Curran Associates, Inc., 2016.
\newblock URL \url{https://proceedings.neurips.cc/paper_files/paper/2016/file/2f885d0fbe2e131bfc9d98363e55d1d4-Paper.pdf}.

\bibitem[OpenAI et~al.(2024)OpenAI, Achiam, Adler, Agarwal, Ahmad, Akkaya, Aleman, Almeida, Altenschmidt, Altman, Anadkat, Avila, Babuschkin, Balaji, Balcom, Baltescu, Bao, Bavarian, Belgum, Bello, Berdine, Bernadett-Shapiro, Berner, Bogdonoff, Boiko, Boyd, Brakman, Brockman, Brooks, Brundage, Button, Cai, Campbell, Cann, Carey, Carlson, Carmichael, Chan, Chang, Chantzis, Chen, Chen, Chen, Chen, Chen, Chess, Cho, Chu, Chung, Cummings, Currier, Dai, Decareaux, Degry, Deutsch, Deville, Dhar, Dohan, Dowling, Dunning, Ecoffet, Eleti, Eloundou, Farhi, Fedus, Felix, Fishman, Forte, Fulford, Gao, Georges, Gibson, Goel, Gogineni, Goh, Gontijo-Lopes, Gordon, Grafstein, Gray, Greene, Gross, Gu, Guo, Hallacy, Han, Harris, He, Heaton, Heidecke, Hesse, Hickey, Hickey, Hoeschele, Houghton, Hsu, Hu, Hu, Huizinga, Jain, Jain, Jang, Jiang, Jiang, Jin, Jin, Jomoto, Jonn, Jun, Kaftan, Łukasz Kaiser, Kamali, Kanitscheider, Keskar, Khan, Kilpatrick, Kim, Kim, Kim, Kirchner, Kiros, Knight, Kokotajlo, Łukasz Kondraciuk, Kondrich,
  Konstantinidis, Kosic, Krueger, Kuo, Lampe, Lan, Lee, Leike, Leung, Levy, Li, Lim, Lin, Lin, Litwin, Lopez, Lowe, Lue, Makanju, Malfacini, Manning, Markov, Markovski, Martin, Mayer, Mayne, McGrew, McKinney, McLeavey, McMillan, McNeil, Medina, Mehta, Menick, Metz, Mishchenko, Mishkin, Monaco, Morikawa, Mossing, Mu, Murati, Murk, Mély, Nair, Nakano, Nayak, Neelakantan, Ngo, Noh, Ouyang, O'Keefe, Pachocki, Paino, Palermo, Pantuliano, Parascandolo, Parish, Parparita, Passos, Pavlov, Peng, Perelman, de~Avila Belbute~Peres, Petrov, de~Oliveira~Pinto, Michael, Pokorny, Pokrass, Pong, Powell, Power, Power, Proehl, Puri, Radford, Rae, Ramesh, Raymond, Real, Rimbach, Ross, Rotsted, Roussez, Ryder, Saltarelli, Sanders, Santurkar, Sastry, Schmidt, Schnurr, Schulman, Selsam, Sheppard, Sherbakov, Shieh, Shoker, Shyam, Sidor, Sigler, Simens, Sitkin, Slama, Sohl, Sokolowsky, Song, Staudacher, Such, Summers, Sutskever, Tang, Tezak, Thompson, Tillet, Tootoonchian, Tseng, Tuggle, Turley, Tworek, Uribe, Vallone, Vijayvergiya,
  Voss, Wainwright, Wang, Wang, Wang, Ward, Wei, Weinmann, Welihinda, Welinder, Weng, Weng, Wiethoff, Willner, Winter, Wolrich, Wong, Workman, Wu, Wu, Wu, Xiao, Xu, Yoo, Yu, Yuan, Zaremba, Zellers, Zhang, Zhang, Zhao, Zheng, Zhuang, Zhuk, and Zoph]{openai2024gpt4}
OpenAI, Josh Achiam, Steven Adler, Sandhini Agarwal, Lama Ahmad, Ilge Akkaya, Florencia~Leoni Aleman, Diogo Almeida, Janko Altenschmidt, Sam Altman, Shyamal Anadkat, Red Avila, Igor Babuschkin, Suchir Balaji, Valerie Balcom, Paul Baltescu, Haiming Bao, Mohammad Bavarian, Jeff Belgum, Irwan Bello, Jake Berdine, Gabriel Bernadett-Shapiro, Christopher Berner, Lenny Bogdonoff, Oleg Boiko, Madelaine Boyd, Anna-Luisa Brakman, Greg Brockman, Tim Brooks, Miles Brundage, Kevin Button, Trevor Cai, Rosie Campbell, Andrew Cann, Brittany Carey, Chelsea Carlson, Rory Carmichael, Brooke Chan, Che Chang, Fotis Chantzis, Derek Chen, Sully Chen, Ruby Chen, Jason Chen, Mark Chen, Ben Chess, Chester Cho, Casey Chu, Hyung~Won Chung, Dave Cummings, Jeremiah Currier, Yunxing Dai, Cory Decareaux, Thomas Degry, Noah Deutsch, Damien Deville, Arka Dhar, David Dohan, Steve Dowling, Sheila Dunning, Adrien Ecoffet, Atty Eleti, Tyna Eloundou, David Farhi, Liam Fedus, Niko Felix, Simón~Posada Fishman, Juston Forte, Isabella Fulford, Leo
  Gao, Elie Georges, Christian Gibson, Vik Goel, Tarun Gogineni, Gabriel Goh, Rapha Gontijo-Lopes, Jonathan Gordon, Morgan Grafstein, Scott Gray, Ryan Greene, Joshua Gross, Shixiang~Shane Gu, Yufei Guo, Chris Hallacy, Jesse Han, Jeff Harris, Yuchen He, Mike Heaton, Johannes Heidecke, Chris Hesse, Alan Hickey, Wade Hickey, Peter Hoeschele, Brandon Houghton, Kenny Hsu, Shengli Hu, Xin Hu, Joost Huizinga, Shantanu Jain, Shawn Jain, Joanne Jang, Angela Jiang, Roger Jiang, Haozhun Jin, Denny Jin, Shino Jomoto, Billie Jonn, Heewoo Jun, Tomer Kaftan, Łukasz Kaiser, Ali Kamali, Ingmar Kanitscheider, Nitish~Shirish Keskar, Tabarak Khan, Logan Kilpatrick, Jong~Wook Kim, Christina Kim, Yongjik Kim, Jan~Hendrik Kirchner, Jamie Kiros, Matt Knight, Daniel Kokotajlo, Łukasz Kondraciuk, Andrew Kondrich, Aris Konstantinidis, Kyle Kosic, Gretchen Krueger, Vishal Kuo, Michael Lampe, Ikai Lan, Teddy Lee, Jan Leike, Jade Leung, Daniel Levy, Chak~Ming Li, Rachel Lim, Molly Lin, Stephanie Lin, Mateusz Litwin, Theresa Lopez, Ryan
  Lowe, Patricia Lue, Anna Makanju, Kim Malfacini, Sam Manning, Todor Markov, Yaniv Markovski, Bianca Martin, Katie Mayer, Andrew Mayne, Bob McGrew, Scott~Mayer McKinney, Christine McLeavey, Paul McMillan, Jake McNeil, David Medina, Aalok Mehta, Jacob Menick, Luke Metz, Andrey Mishchenko, Pamela Mishkin, Vinnie Monaco, Evan Morikawa, Daniel Mossing, Tong Mu, Mira Murati, Oleg Murk, David Mély, Ashvin Nair, Reiichiro Nakano, Rajeev Nayak, Arvind Neelakantan, Richard Ngo, Hyeonwoo Noh, Long Ouyang, Cullen O'Keefe, Jakub Pachocki, Alex Paino, Joe Palermo, Ashley Pantuliano, Giambattista Parascandolo, Joel Parish, Emy Parparita, Alex Passos, Mikhail Pavlov, Andrew Peng, Adam Perelman, Filipe de~Avila Belbute~Peres, Michael Petrov, Henrique~Ponde de~Oliveira~Pinto, Michael, Pokorny, Michelle Pokrass, Vitchyr~H. Pong, Tolly Powell, Alethea Power, Boris Power, Elizabeth Proehl, Raul Puri, Alec Radford, Jack Rae, Aditya Ramesh, Cameron Raymond, Francis Real, Kendra Rimbach, Carl Ross, Bob Rotsted, Henri Roussez,
  Nick Ryder, Mario Saltarelli, Ted Sanders, Shibani Santurkar, Girish Sastry, Heather Schmidt, David Schnurr, John Schulman, Daniel Selsam, Kyla Sheppard, Toki Sherbakov, Jessica Shieh, Sarah Shoker, Pranav Shyam, Szymon Sidor, Eric Sigler, Maddie Simens, Jordan Sitkin, Katarina Slama, Ian Sohl, Benjamin Sokolowsky, Yang Song, Natalie Staudacher, Felipe~Petroski Such, Natalie Summers, Ilya Sutskever, Jie Tang, Nikolas Tezak, Madeleine~B. Thompson, Phil Tillet, Amin Tootoonchian, Elizabeth Tseng, Preston Tuggle, Nick Turley, Jerry Tworek, Juan Felipe~Cerón Uribe, Andrea Vallone, Arun Vijayvergiya, Chelsea Voss, Carroll Wainwright, Justin~Jay Wang, Alvin Wang, Ben Wang, Jonathan Ward, Jason Wei, CJ~Weinmann, Akila Welihinda, Peter Welinder, Jiayi Weng, Lilian Weng, Matt Wiethoff, Dave Willner, Clemens Winter, Samuel Wolrich, Hannah Wong, Lauren Workman, Sherwin Wu, Jeff Wu, Michael Wu, Kai Xiao, Tao Xu, Sarah Yoo, Kevin Yu, Qiming Yuan, Wojciech Zaremba, Rowan Zellers, Chong Zhang, Marvin Zhang, Shengjia
  Zhao, Tianhao Zheng, Juntang Zhuang, William Zhuk, and Barret Zoph.
\newblock Gpt-4 technical report, 2024.

\bibitem[Otwinowski et~al.(2020)Otwinowski, LaMont, and Nourmohammad]{e22090967}
Jakub Otwinowski, Colin~H. LaMont, and Armita Nourmohammad.
\newblock Information-geometric optimization with natural selection.
\newblock \emph{Entropy}, 22\penalty0 (9), 2020.
\newblock ISSN 1099-4300.
\newblock \doi{10.3390/e22090967}.
\newblock URL \url{https://www.mdpi.com/1099-4300/22/9/967}.

\bibitem[Ouyang et~al.(2022)Ouyang, Wu, Jiang, Almeida, Wainwright, Mishkin, Zhang, Agarwal, Slama, Gray, Schulman, Hilton, Kelton, Miller, Simens, Askell, Welinder, Christiano, Leike, and Lowe]{ouyang2022training}
Long Ouyang, Jeffrey Wu, Xu~Jiang, Diogo Almeida, Carroll Wainwright, Pamela Mishkin, Chong Zhang, Sandhini Agarwal, Katarina Slama, Alex Gray, John Schulman, Jacob Hilton, Fraser Kelton, Luke Miller, Maddie Simens, Amanda Askell, Peter Welinder, Paul Christiano, Jan Leike, and Ryan Lowe.
\newblock Training language models to follow instructions with human feedback.
\newblock In Alice~H. Oh, Alekh Agarwal, Danielle Belgrave, and Kyunghyun Cho (eds.), \emph{Advances in Neural Information Processing Systems}, 2022.
\newblock URL \url{https://openreview.net/forum?id=TG8KACxEON}.

\bibitem[Ouyang et~al.(2024)Ouyang, Zhang, Yan, Liu, Choi, Han, and Qin]{ouyang2024structured}
Siru Ouyang, Zhuosheng Zhang, Bing Yan, Xuan Liu, Yejin Choi, Jiawei Han, and Lianhui Qin.
\newblock Structured chemistry reasoning with large language models, 2024.

\bibitem[Padmakumar et~al.(2023)Padmakumar, Pang, He, and Parikh]{padmakumar2023extrapolative}
Vishakh Padmakumar, Richard~Yuanzhe Pang, He~He, and Ankur~P Parikh.
\newblock Extrapolative controlled sequence generation via iterative refinement.
\newblock \emph{Fortieth International Conference on Machine Learning (ICML)}, 2023.

\bibitem[Peng et~al.(2023)Peng, Galley, He, Cheng, Xie, Hu, Huang, Liden, Yu, Chen, and Gao]{peng2023check}
Baolin Peng, Michel Galley, Pengcheng He, Hao Cheng, Yujia Xie, Yu~Hu, Qiuyuan Huang, Lars Liden, Zhou Yu, Weizhu Chen, and Jianfeng Gao.
\newblock Check your facts and try again: Improving large language models with external knowledge and automated feedback, 2023.

\bibitem[Prabhumoye et~al.(2018)Prabhumoye, Tsvetkov, Salakhutdinov, and Black]{prabhumoye-etal-2018-style}
Shrimai Prabhumoye, Yulia Tsvetkov, Ruslan Salakhutdinov, and Alan~W Black.
\newblock Style transfer through back-translation.
\newblock In Iryna Gurevych and Yusuke Miyao (eds.), \emph{Proceedings of the 56th Annual Meeting of the Association for Computational Linguistics (Volume 1: Long Papers)}, pp.\  866--876, Melbourne, Australia, July 2018. Association for Computational Linguistics.
\newblock \doi{10.18653/v1/P18-1080}.
\newblock URL \url{https://aclanthology.org/P18-1080}.

\bibitem[Qin et~al.(2022)Qin, Welleck, Khashabi, and Choi]{NEURIPS2022_3e25d1af}
Lianhui Qin, Sean Welleck, Daniel Khashabi, and Yejin Choi.
\newblock Cold decoding: Energy-based constrained text generation with langevin dynamics.
\newblock In S.~Koyejo, S.~Mohamed, A.~Agarwal, D.~Belgrave, K.~Cho, and A.~Oh (eds.), \emph{Advances in Neural Information Processing Systems}, volume~35, pp.\  9538--9551. Curran Associates, Inc., 2022.
\newblock URL \url{https://proceedings.neurips.cc/paper_files/paper/2022/file/3e25d1aff47964c8409fd5c8dc0438d7-Paper-Conference.pdf}.

\bibitem[Rabbani et~al.(2023)Rabbani, Ahmad, Ahmad, and Khan]{RABBANI2023822}
Gulam Rabbani, Ejaz Ahmad, Abrar Ahmad, and Rizwan~Hasan Khan.
\newblock Structural features, temperature adaptation and industrial applications of microbial lipases from psychrophilic, mesophilic and thermophilic origins.
\newblock \emph{International Journal of Biological Macromolecules}, 225:\penalty0 822--839, 2023.
\newblock ISSN 0141-8130.
\newblock \doi{https://doi.org/10.1016/j.ijbiomac.2022.11.146}.
\newblock URL \url{https://www.sciencedirect.com/science/article/pii/S0141813022027076}.

\bibitem[Ramachandran et~al.(2022)Ramachandran, Hashimoto, and Xiong]{ramachandran-etal-2022-caspi}
Govardana~Sachithanandam Ramachandran, Kazuma Hashimoto, and Caiming Xiong.
\newblock [{CASPI}] causal-aware safe policy improvement for task-oriented dialogue.
\newblock In Smaranda Muresan, Preslav Nakov, and Aline Villavicencio (eds.), \emph{Proceedings of the 60th Annual Meeting of the Association for Computational Linguistics (Volume 1: Long Papers)}, pp.\  92--102, Dublin, Ireland, May 2022. Association for Computational Linguistics.
\newblock \doi{10.18653/v1/2022.acl-long.8}.
\newblock URL \url{https://aclanthology.org/2022.acl-long.8}.

\bibitem[Ren et~al.(2022)Ren, Li, Ding, Zhou, Ma, and Peng]{pmlr-v162-ren22a}
Zhizhou Ren, Jiahan Li, Fan Ding, Yuan Zhou, Jianzhu Ma, and Jian Peng.
\newblock Proximal exploration for model-guided protein sequence design.
\newblock In Kamalika Chaudhuri, Stefanie Jegelka, Le~Song, Csaba Szepesvari, Gang Niu, and Sivan Sabato (eds.), \emph{Proceedings of the 39th International Conference on Machine Learning}, volume 162 of \emph{Proceedings of Machine Learning Research}, pp.\  18520--18536. PMLR, 17--23 Jul 2022.
\newblock URL \url{https://proceedings.mlr.press/v162/ren22a.html}.

\bibitem[Riley et~al.(2021)Riley, Constant, Guo, Kumar, Uthus, and Parekh]{riley-etal-2021-textsettr}
Parker Riley, Noah Constant, Mandy Guo, Girish Kumar, David Uthus, and Zarana Parekh.
\newblock {T}ext{SETTR}: Few-shot text style extraction and tunable targeted restyling.
\newblock In Chengqing Zong, Fei Xia, Wenjie Li, and Roberto Navigli (eds.), \emph{Proceedings of the 59th Annual Meeting of the Association for Computational Linguistics and the 11th International Joint Conference on Natural Language Processing (Volume 1: Long Papers)}, pp.\  3786--3800, Online, August 2021. Association for Computational Linguistics.
\newblock \doi{10.18653/v1/2021.acl-long.293}.
\newblock URL \url{https://aclanthology.org/2021.acl-long.293}.

\bibitem[Russo et~al.(2020)Russo, Hollenstein, Musat, and Zhang]{russo-etal-2020-control}
Giuseppe Russo, Nora Hollenstein, Claudiu~Cristian Musat, and Ce~Zhang.
\newblock Control, generate, augment: A scalable framework for multi-attribute text generation.
\newblock In Trevor Cohn, Yulan He, and Yang Liu (eds.), \emph{Findings of the Association for Computational Linguistics: EMNLP 2020}, pp.\  351--366, Online, November 2020. Association for Computational Linguistics.
\newblock \doi{10.18653/v1/2020.findings-emnlp.33}.
\newblock URL \url{https://aclanthology.org/2020.findings-emnlp.33}.

\bibitem[Sarkisyan et~al.(2016)Sarkisyan, Bolotin, Meer, Usmanova, Mishin, Sharonov, Ivankov, Bozhanova, Baranov, Soylemez, et~al.]{sarkisyan2016local}
Karen~S Sarkisyan, Dmitry~A Bolotin, Margarita~V Meer, Dinara~R Usmanova, Alexander~S Mishin, George~V Sharonov, Dmitry~N Ivankov, Nina~G Bozhanova, Mikhail~S Baranov, Onuralp Soylemez, et~al.
\newblock Local fitness landscape of the green fluorescent protein.
\newblock \emph{Nature}, 533\penalty0 (7603):\penalty0 397--401, 2016.

\bibitem[Schlinkmann et~al.(2012)Schlinkmann, Honegger, Türeci, Robison, Lipovšek, and Plückthun]{Schlinkmann2012CriticalFF}
Karola~M. Schlinkmann, Annemarie Honegger, Esin Türeci, Keith~E. Robison, Daša Lipovšek, and Andreas Plückthun.
\newblock Critical features for biosynthesis, stability, and functionality of a g protein-coupled receptor uncovered by all-versus-all mutations.
\newblock \emph{Proceedings of the National Academy of Sciences}, 109\penalty0 (25):\penalty0 9810--9815, 2012.
\newblock \doi{10.1073/pnas.1202107109}.
\newblock URL \url{https://www.pnas.org/doi/abs/10.1073/pnas.1202107109}.

\bibitem[Schymkowitz et~al.(2005)Schymkowitz, Borg, Stricher, Nys, Rousseau, and Serrano]{schymkowitz2005foldx}
Joost Schymkowitz, Jesper Borg, Francois Stricher, Robby Nys, Frederic Rousseau, and Luis Serrano.
\newblock The foldx web server: an online force field.
\newblock \emph{Nucleic acids research}, 33\penalty0 (suppl\_2):\penalty0 W382--W388, 2005.

\bibitem[Shaner et~al.(2007)Shaner, Patterson, and Davidson]{10.1242_jcs.005801}
Nathan~C. Shaner, George~H. Patterson, and Michael~W. Davidson.
\newblock {Advances in fluorescent protein technology}.
\newblock \emph{Journal of Cell Science}, 120\penalty0 (24):\penalty0 4247--4260, 12 2007.
\newblock ISSN 0021-9533.
\newblock \doi{10.1242/jcs.005801}.
\newblock URL \url{https://doi.org/10.1242/jcs.005801}.

\bibitem[Shen et~al.(2014)Shen, Wong, Xiao, Guo, and Smale]{shen2014introduction}
Wen-Jun Shen, Hau-San Wong, Quan-Wu Xiao, Xin Guo, and Stephen Smale.
\newblock Introduction to the peptide binding problem of computational immunology: new results.
\newblock \emph{Foundations of Computational Mathematics}, 14:\penalty0 951--984, 2014.

\bibitem[Shinn et~al.(2023)Shinn, Cassano, Gopinath, Narasimhan, and Yao]{shinn2023reflexion}
Noah Shinn, Federico Cassano, Ashwin Gopinath, Karthik~R Narasimhan, and Shunyu Yao.
\newblock Reflexion: language agents with verbal reinforcement learning.
\newblock In \emph{Thirty-seventh Conference on Neural Information Processing Systems}, 2023.
\newblock URL \url{https://openreview.net/forum?id=vAElhFcKW6}.

\bibitem[Sinai et~al.(2020)Sinai, Wang, Whatley, Slocum, Locane, and Kelsic]{sinai2020adalead}
Sam Sinai, Richard Wang, Alexander Whatley, Stewart Slocum, Elina Locane, and Eric~D. Kelsic.
\newblock Adalead: A simple and robust adaptive greedy search algorithm for sequence design, 2020.

\bibitem[Starr et~al.(2020)Starr, Greaney, Hilton, Ellis, Crawford, Dingens, Navarro, Bowen, Tortorici, Walls, et~al.]{starr2020deep}
Tyler~N Starr, Allison~J Greaney, Sarah~K Hilton, Daniel Ellis, Katharine~HD Crawford, Adam~S Dingens, Mary~Jane Navarro, John~E Bowen, M~Alejandra Tortorici, Alexandra~C Walls, et~al.
\newblock Deep mutational scanning of sars-cov-2 receptor binding domain reveals constraints on folding and ace2 binding.
\newblock \emph{cell}, 182\penalty0 (5):\penalty0 1295--1310, 2020.

\bibitem[Sternke \& Karpiak(2023)Sternke and Karpiak]{sternke2023proteinrl}
Matt Sternke and Joel Karpiak.
\newblock Protein{RL}: Reinforcement learning with generative protein language models for property-directed sequence design.
\newblock In \emph{NeurIPS 2023 Generative AI and Biology (GenBio) Workshop}, 2023.
\newblock URL \url{https://openreview.net/forum?id=sWCsSKqkXa}.

\bibitem[Touvron et~al.(2023)Touvron, Martin, Stone, Albert, Almahairi, Babaei, Bashlykov, Batra, Bhargava, Bhosale, Bikel, Blecher, Ferrer, Chen, Cucurull, Esiobu, Fernandes, Fu, Fu, Fuller, Gao, Goswami, Goyal, Hartshorn, Hosseini, Hou, Inan, Kardas, Kerkez, Khabsa, Kloumann, Korenev, Koura, Lachaux, Lavril, Lee, Liskovich, Lu, Mao, Martinet, Mihaylov, Mishra, Molybog, Nie, Poulton, Reizenstein, Rungta, Saladi, Schelten, Silva, Smith, Subramanian, Tan, Tang, Taylor, Williams, Kuan, Xu, Yan, Zarov, Zhang, Fan, Kambadur, Narang, Rodriguez, Stojnic, Edunov, and Scialom]{touvron2023llama}
Hugo Touvron, Louis Martin, Kevin Stone, Peter Albert, Amjad Almahairi, Yasmine Babaei, Nikolay Bashlykov, Soumya Batra, Prajjwal Bhargava, Shruti Bhosale, Dan Bikel, Lukas Blecher, Cristian~Canton Ferrer, Moya Chen, Guillem Cucurull, David Esiobu, Jude Fernandes, Jeremy Fu, Wenyin Fu, Brian Fuller, Cynthia Gao, Vedanuj Goswami, Naman Goyal, Anthony Hartshorn, Saghar Hosseini, Rui Hou, Hakan Inan, Marcin Kardas, Viktor Kerkez, Madian Khabsa, Isabel Kloumann, Artem Korenev, Punit~Singh Koura, Marie-Anne Lachaux, Thibaut Lavril, Jenya Lee, Diana Liskovich, Yinghai Lu, Yuning Mao, Xavier Martinet, Todor Mihaylov, Pushkar Mishra, Igor Molybog, Yixin Nie, Andrew Poulton, Jeremy Reizenstein, Rashi Rungta, Kalyan Saladi, Alan Schelten, Ruan Silva, Eric~Michael Smith, Ranjan Subramanian, Xiaoqing~Ellen Tan, Binh Tang, Ross Taylor, Adina Williams, Jian~Xiang Kuan, Puxin Xu, Zheng Yan, Iliyan Zarov, Yuchen Zhang, Angela Fan, Melanie Kambadur, Sharan Narang, Aurelien Rodriguez, Robert Stojnic, Sergey Edunov, and Thomas
  Scialom.
\newblock Llama 2: Open foundation and fine-tuned chat models, 2023.

\bibitem[Wang et~al.(2020)Wang, Novikov, Zolna, Merel, Springenberg, Reed, Shahriari, Siegel, Gulcehre, Heess, and de~Freitas]{NEURIPS2020_588cb956}
Ziyu Wang, Alexander Novikov, Konrad Zolna, Josh~S Merel, Jost~Tobias Springenberg, Scott~E Reed, Bobak Shahriari, Noah Siegel, Caglar Gulcehre, Nicolas Heess, and Nando de~Freitas.
\newblock Critic regularized regression.
\newblock In H.~Larochelle, M.~Ranzato, R.~Hadsell, M.F. Balcan, and H.~Lin (eds.), \emph{Advances in Neural Information Processing Systems}, volume~33, pp.\  7768--7778. Curran Associates, Inc., 2020.
\newblock URL \url{https://proceedings.neurips.cc/paper_files/paper/2020/file/588cb956d6bbe67078f29f8de420a13d-Paper.pdf}.

\bibitem[Warstadt et~al.(2019)Warstadt, Singh, and Bowman]{warstadt-etal-2019-neural}
Alex Warstadt, Amanpreet Singh, and Samuel~R. Bowman.
\newblock Neural network acceptability judgments.
\newblock \emph{Transactions of the Association for Computational Linguistics}, 7:\penalty0 625--641, 2019.
\newblock \doi{10.1162/tacl_a_00290}.
\newblock URL \url{https://aclanthology.org/Q19-1040}.

\bibitem[Welleck et~al.(2023)Welleck, Lu, West, Brahman, Shen, Khashabi, and Choi]{welleck2023generating}
Sean Welleck, Ximing Lu, Peter West, Faeze Brahman, Tianxiao Shen, Daniel Khashabi, and Yejin Choi.
\newblock Generating sequences by learning to self-correct.
\newblock In \emph{The Eleventh International Conference on Learning Representations}, 2023.
\newblock URL \url{https://openreview.net/forum?id=hH36JeQZDaO}.

\bibitem[Whitehead et~al.(2012)Whitehead, Chevalier, Song, Dreyfus, Fleishman, De~Mattos, Myers, Kamisetty, Blair, Wilson, et~al.]{whitehead2012optimization}
Timothy~A Whitehead, Aaron Chevalier, Yifan Song, Cyrille Dreyfus, Sarel~J Fleishman, Cecilia De~Mattos, Chris~A Myers, Hetunandan Kamisetty, Patrick Blair, Ian~A Wilson, et~al.
\newblock Optimization of affinity, specificity and function of designed influenza inhibitors using deep sequencing.
\newblock \emph{Nature biotechnology}, 30\penalty0 (6):\penalty0 543--548, 2012.

\bibitem[Wong et~al.(2023)Wong, Zheng, Valeri, Donghia, Anahtar, Omori, Li, Cubillos-Ruiz, Krishnan, Jin, et~al.]{wong2023discovery}
Felix Wong, Erica~J Zheng, Jacqueline~A Valeri, Nina~M Donghia, Melis~N Anahtar, Satotaka Omori, Alicia Li, Andres Cubillos-Ruiz, Aarti Krishnan, Wengong Jin, et~al.
\newblock Discovery of a structural class of antibiotics with explainable deep learning.
\newblock \emph{Nature}, pp.\  1--9, 2023.

\bibitem[Xu et~al.(2018)Xu, Sun, Zeng, Zhang, Ren, Wang, and Li]{xu-etal-2018-unpaired}
Jingjing Xu, Xu~Sun, Qi~Zeng, Xiaodong Zhang, Xuancheng Ren, Houfeng Wang, and Wenjie Li.
\newblock Unpaired sentiment-to-sentiment translation: A cycled reinforcement learning approach.
\newblock In Iryna Gurevych and Yusuke Miyao (eds.), \emph{Proceedings of the 56th Annual Meeting of the Association for Computational Linguistics (Volume 1: Long Papers)}, pp.\  979--988, Melbourne, Australia, July 2018. Association for Computational Linguistics.
\newblock \doi{10.18653/v1/P18-1090}.
\newblock URL \url{https://aclanthology.org/P18-1090}.

\bibitem[Yang et~al.(2023)Yang, Zhang, Xia, Feng, Xiong, and Zhou]{yang2023preferencegrounded}
Shentao Yang, Shujian Zhang, Congying Xia, Yihao Feng, Caiming Xiong, and Mingyuan Zhou.
\newblock Preference-grounded token-level guidance for language model fine-tuning, 2023.

\bibitem[Zhang et~al.(2023)Zhang, Liu, and Zhang]{zhang-etal-2023-summit}
Haopeng Zhang, Xiao Liu, and Jiawei Zhang.
\newblock {S}umm{I}t: Iterative text summarization via {C}hat{GPT}.
\newblock In Houda Bouamor, Juan Pino, and Kalika Bali (eds.), \emph{Findings of the Association for Computational Linguistics: EMNLP 2023}, pp.\  10644--10657, Singapore, December 2023. Association for Computational Linguistics.
\newblock \doi{10.18653/v1/2023.findings-emnlp.714}.
\newblock URL \url{https://aclanthology.org/2023.findings-emnlp.714}.

\bibitem[Zhang et~al.(2024)Zhang, Zeng, Wang, and Lu]{zhang2024tinyllama}
Peiyuan Zhang, Guangtao Zeng, Tianduo Wang, and Wei Lu.
\newblock Tinyllama: An open-source small language model, 2024.

\bibitem[Zhang et~al.(2015)Zhang, Zhao, and LeCun]{NIPS2015_250cf8b5}
Xiang Zhang, Junbo Zhao, and Yann LeCun.
\newblock Character-level convolutional networks for text classification.
\newblock In C.~Cortes, N.~Lawrence, D.~Lee, M.~Sugiyama, and R.~Garnett (eds.), \emph{Advances in Neural Information Processing Systems}, volume~28. Curran Associates, Inc., 2015.
\newblock URL \url{https://proceedings.neurips.cc/paper_files/paper/2015/file/250cf8b51c773f3f8dc8b4be867a9a02-Paper.pdf}.

\bibitem[Zhang et~al.(2018)Zhang, Ren, Liu, Wang, Chen, Li, Zhou, and Chen]{zhang2018style}
Zhirui Zhang, Shuo Ren, Shujie Liu, Jianyong Wang, Peng Chen, Mu~Li, Ming Zhou, and Enhong Chen.
\newblock Style transfer as unsupervised machine translation, 2018.

\end{thebibliography}
\bibliographystyle{tmlr}

\appendix

\section{{\benchmark} - Text Style Transfer Implementation Details}

\subsection{Sentiment and Complexity Regressor Training}
\label{sec:sentiment_complexity_regressor}
\paragraph{Sentiment Regressor} We train Sentiment regressor on Yelp reviews \citep{NIPS2015_250cf8b5}. The original data contained $650K$ train and $50K$ test reviews divided evenly across five labels (1 - very negative, 2 - negative, 3 - neutral, 4 - positive, and 5 - very positive). We filter reviews that are non-English\footnote{Using external language identification classifier \url{https://huggingface.co/papluca/xlm-roberta-base-language-detection}.} or long ($>200$ tokens). After filtering, we obtain $\approx 464K$ train reviews and $\approx 36K$ test reviews. We randomly sample $\approx 36K$ reviews from the train set for validation and train a RoBERTa-large \citep{liu2020roberta} regressor on the remaining instances for 4 epochs using mean squared error loss. The final regressor obtained a $0.92$ test correlation (and $0.37$ mean absolute error). During inference, we clamp the predictions from the regressor such that its output range is $\in [1, 5]$.

\paragraph{Complexity Regressor} To obtain the Complexity regressor we train a ranking model on top of the SWiPE Wikipedia simplification dataset \citep{laban-etal-2023-swipe}. The SWiPE dataset contains $\approx 143K$ pairs of simple to complex Wikipedia paragraphs. However, many instances were low quality (very long, very short, high repetition, bad words, non-English, etc.). After filtering these instances, we are left with $79K$ train, $1K$ validation, and $1.8K$ test simple to complex pairs. We train a RoBERTa-large \citep{liu2020roberta} regressor on this pairwise data using the ranking objective \citep{10.1162/0891201053630273} for 8 epochs. The best checkpoint emitted raw scores in the range of $\in [-17.1, 17.1]$ and obtained $98.2\%$ accuracy on the test set (comparing the raw scores of simple and complex passage pairs). We linearly interpolate this output range to be $\in [-2, 2]$ such that we can subdivide the output range from the regressor into five fine-grained threshold boundaries (to match the Sentiment regressor labels).

\begin{figure}[t]
\centering
\includegraphics[width=\linewidth]{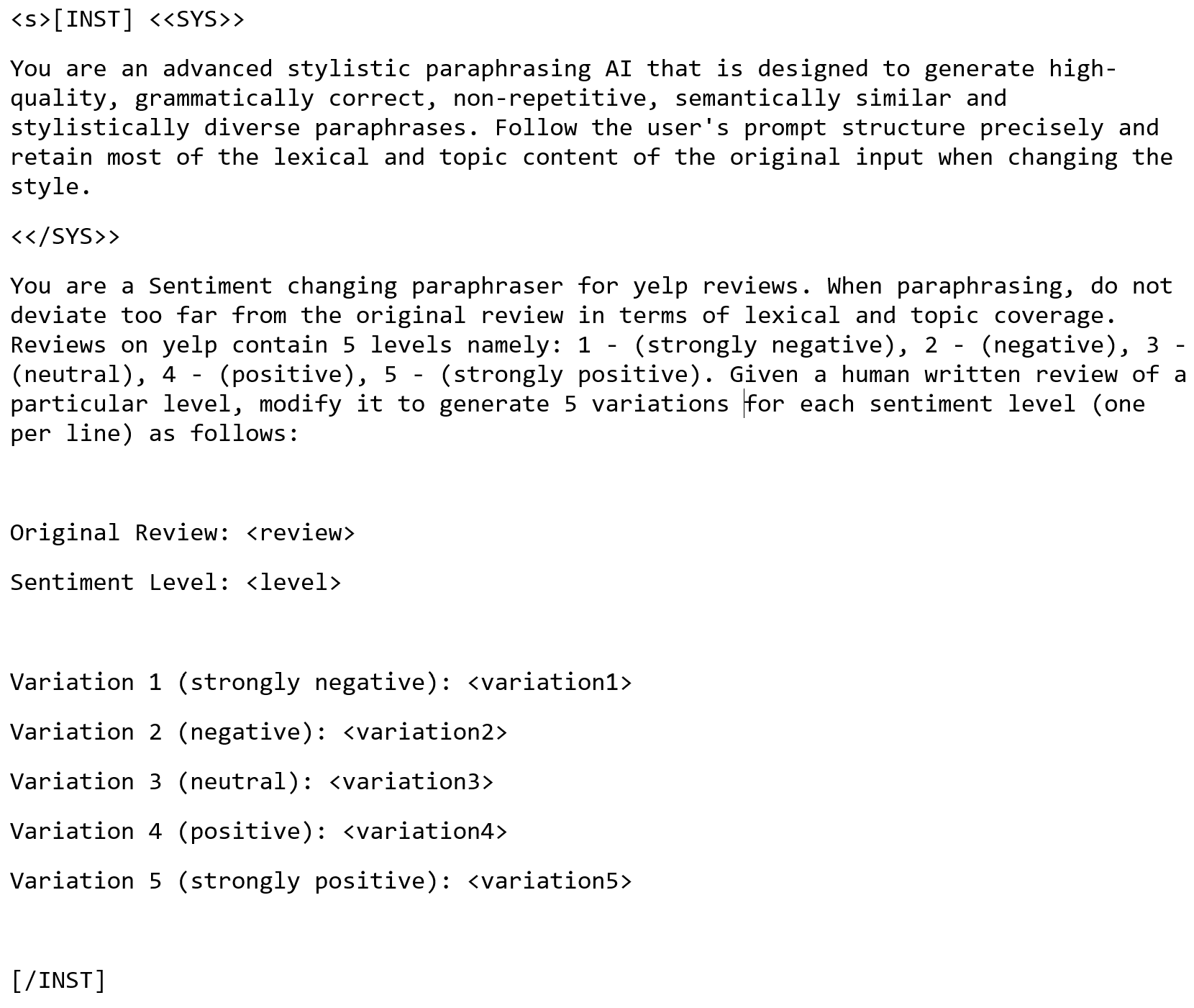}
\caption{Llama2 Sentiment paraphrasing prompt}
\label{fig:sentiment_prompt}
\end{figure}

\begin{figure}[t]
\centering
\includegraphics[width=\linewidth]{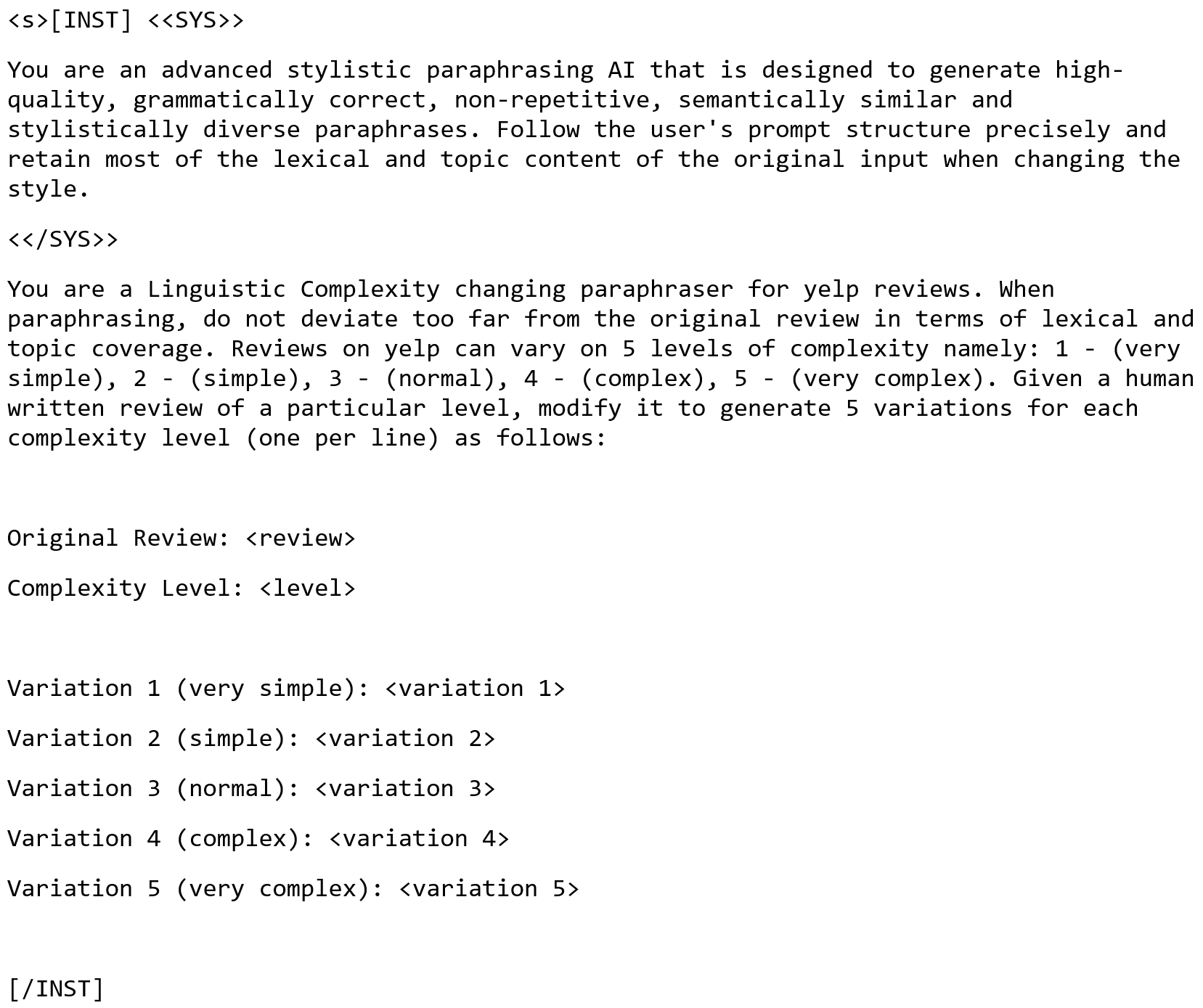}
\caption{Llama2 Complexity paraphrasing prompt}
\label{fig:complexity_prompt}
\end{figure}

\subsection{Sentiment and Complexity Few-Shot Prompts}
\label{sec:few_shot_variations}
To generate attributed variations of Yelp reviews in the Sentiment and Complexity axis we use few-shot prompting on top of a Llama2-7B\footnote{\href{https://huggingface.co/meta-llama/Llama-2-7b-chat-hf}{meta-llama/Llama-2-7b-chat-hf}} parameter model \citep{touvron2023llama}. The prompt used for Sentiment and Complexity are given in Figures \ref{fig:sentiment_prompt} and \ref{fig:complexity_prompt} respectively. We augment both prompts with their own 3-shot demonstrations and generate 5 samples for each review using nucleus sampling ($top_p = 0.95$).

\section{{\benchmark} - Protein Design Implementation Details}

\subsection{Fluorescence and ddG Regressor Training}
\label{subsec:protein_regressor_impl}
To train the protein evaluators, we randomly split the $\approx 51.7K$ mutant sequences into 50\% train, 15\% validation, and 35\% test sequences. Along with the ESM2 model-based regressors, we also experimented with traditional CNN regressors \citep{dallago2021flip}. The learning rate for both models is 1e-4 where the ESM2-based regressor is trained for 12 epochs and the CNN regressor was trained for 40 epochs. Despite additional training time, the test set correlation for fluorescence and ddG for the CNN regressors are $0.892$ and $0.933$ respectively, that are much lower than ESM2-based models ($0.974$ and $0.987$ respectively). Subsequently, we use the ESM2-based regressors as evaluators in our experiments.

\begin{figure}[t]
\begin{subfigure}[t]{0.49\textwidth}
  \centering
  \includegraphics[width=\linewidth]{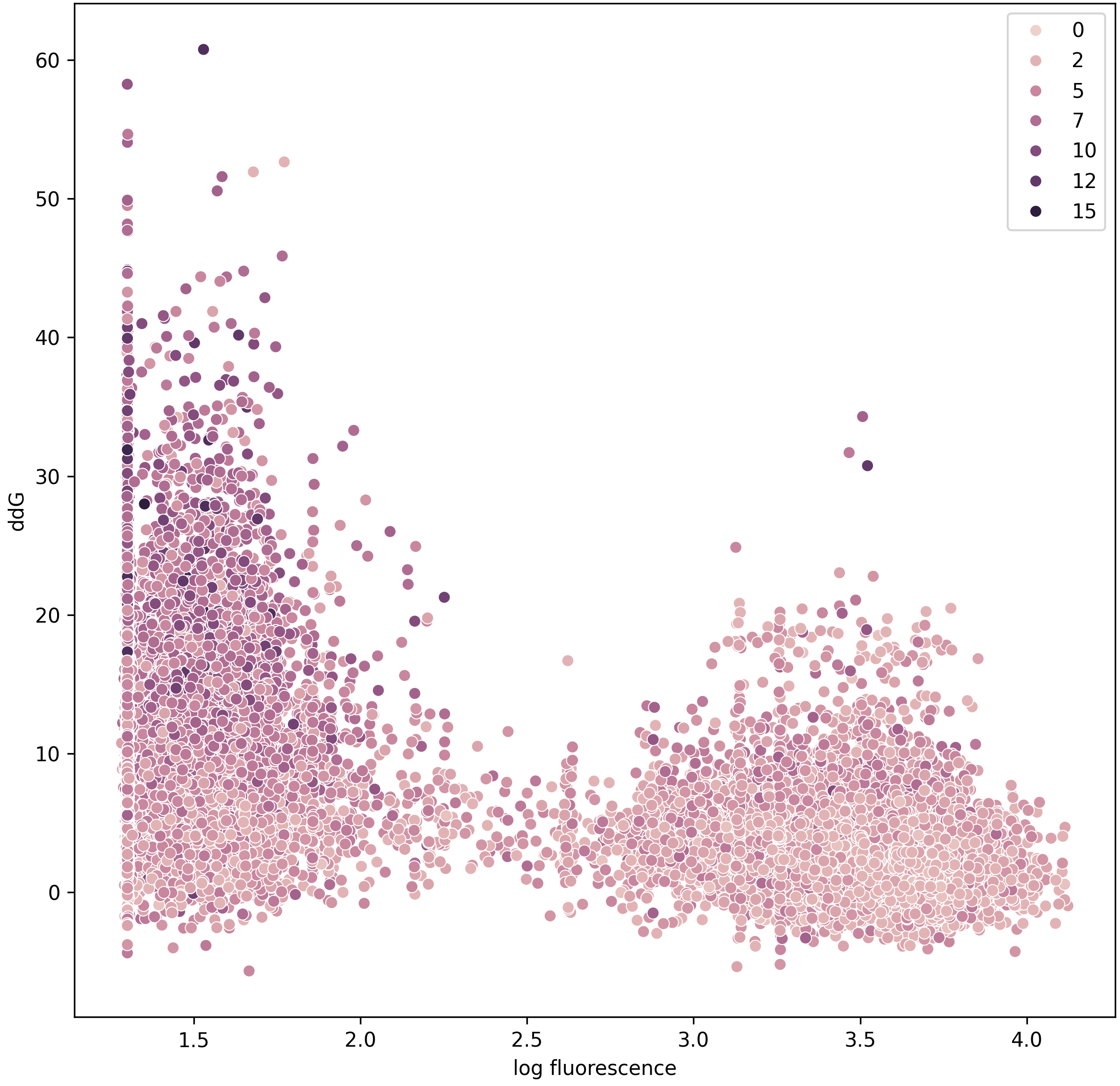}
  \caption{Mutant edit distances from wild-type GFP} 
  \label{fig:edit_distance_dist}
\end{subfigure}\hfill
\begin{subfigure}[t]{0.49\textwidth}
  \centering
  \includegraphics[width=\linewidth]{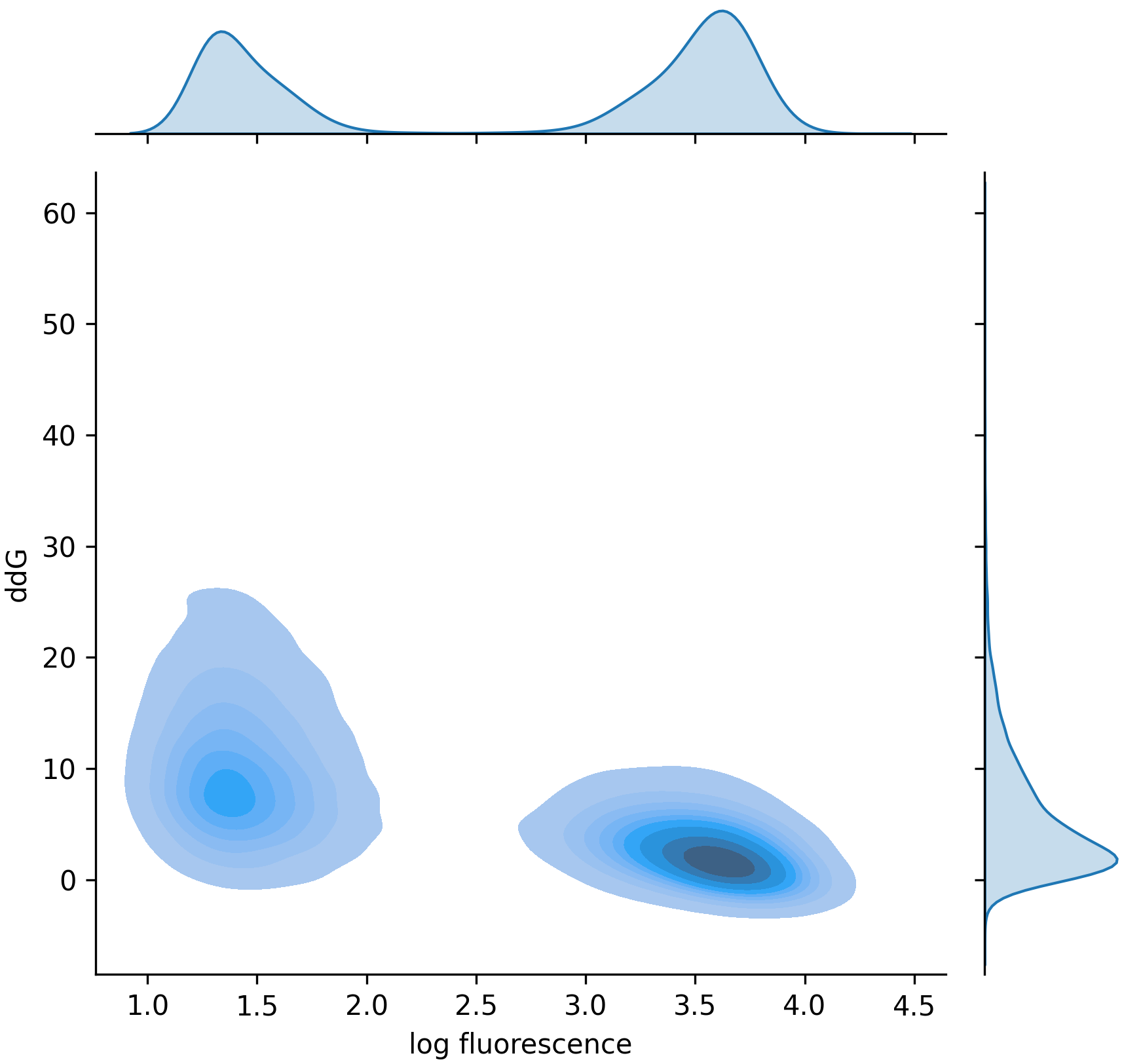}
  \caption{Attribute distribution of GFP mutants} 
  \label{fig:mutant_density}
\end{subfigure}
\caption{Log Fluorescence and ddG distribution of $51.6K$ GFP mutants}
\label{fig:GFP_dist}
\end{figure}

\subsection{Protein Design Baselines and LM editor Implementation}
\label{subsec:protein_editors_impl}
The $51.7K$ GFP train mutants are unevenly divided across the 16 multi-attribute threshold combinations as seen in Figure \ref{fig:mut_dist}. In the Random mutation baseline, when predicting new mutant sequences from a particular threshold combination, we maintain the edit distance distribution of the train sequences within the same threshold combination. The Recombine baseline uses a recombination strategy where a pair of sequences are mixed (shuffling each position with a recombination rate $\kappa = 0.5$) to create two new sequences. When generating new sequences for a particular threshold combination with the Recombine baseline, we set the train sequences within the same thresholds as the seed set, randomly shuffle them, and iteratively apply the recombination strategy until we get 3000 new sequences. Since some threshold combinations have very low seed sequences (<200) there may be duplicates when generating the 3000 new sequences with this strategy. We improve upon this baseline, we create Unique Recombine where we keep generating sequences with the recombination strategy until we get 3000 unique sequences that don't overlap with the training set. 

For the Protein Language Model editor models trained with our {\framework} framework, we use the nucleus sampling \citep{holtzman2019curious} with ($top_p = 0.95$). We also had to increase the generation temperature to $1.2$ to encourage more diverse sequences. During our early experiments, we sampled edit pairs from each threshold combination uniformly, leading to overfitting in the low-density regions of multi-attribute space, i.e., most LM-generated sequences are duplicated. To mitigate this behavior, we downsampled the threshold combinations containing fewer than $\tau=400$ sequences.\footnote{If a target threshold combination is $n < \tau$ sequences, we reduce its edit-pair sampling weights to $n/\tau$ to reduce overfitting in the sparse region.}

\begin{figure}[t]
\begin{subfigure}[t]{0.48\textwidth}
  \centering
  \includegraphics[width=\linewidth]{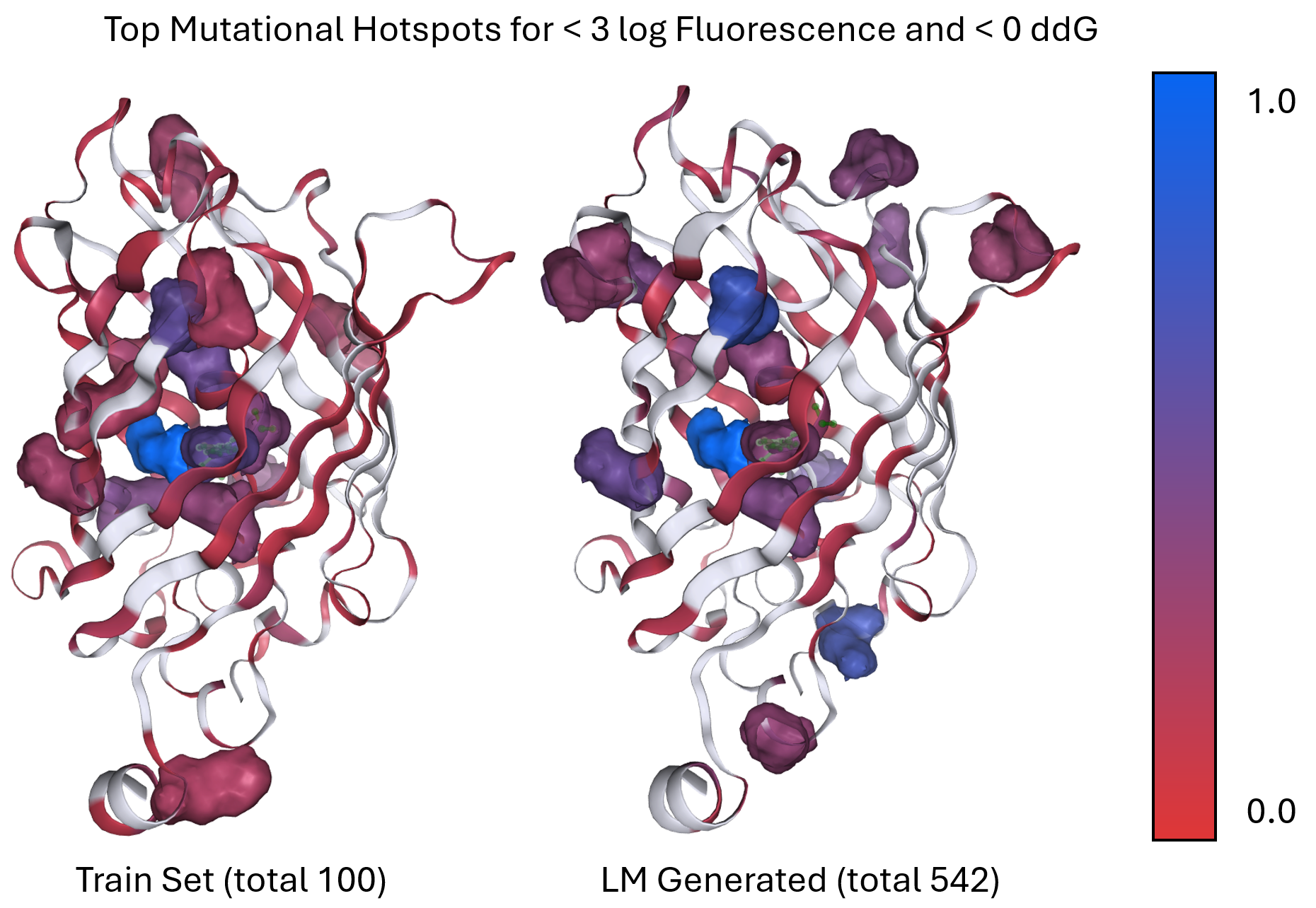}
  \caption{low fluorescence high stability} 
\end{subfigure}
\begin{subfigure}[t]{0.48\textwidth}
  \centering
  \includegraphics[width=\linewidth]{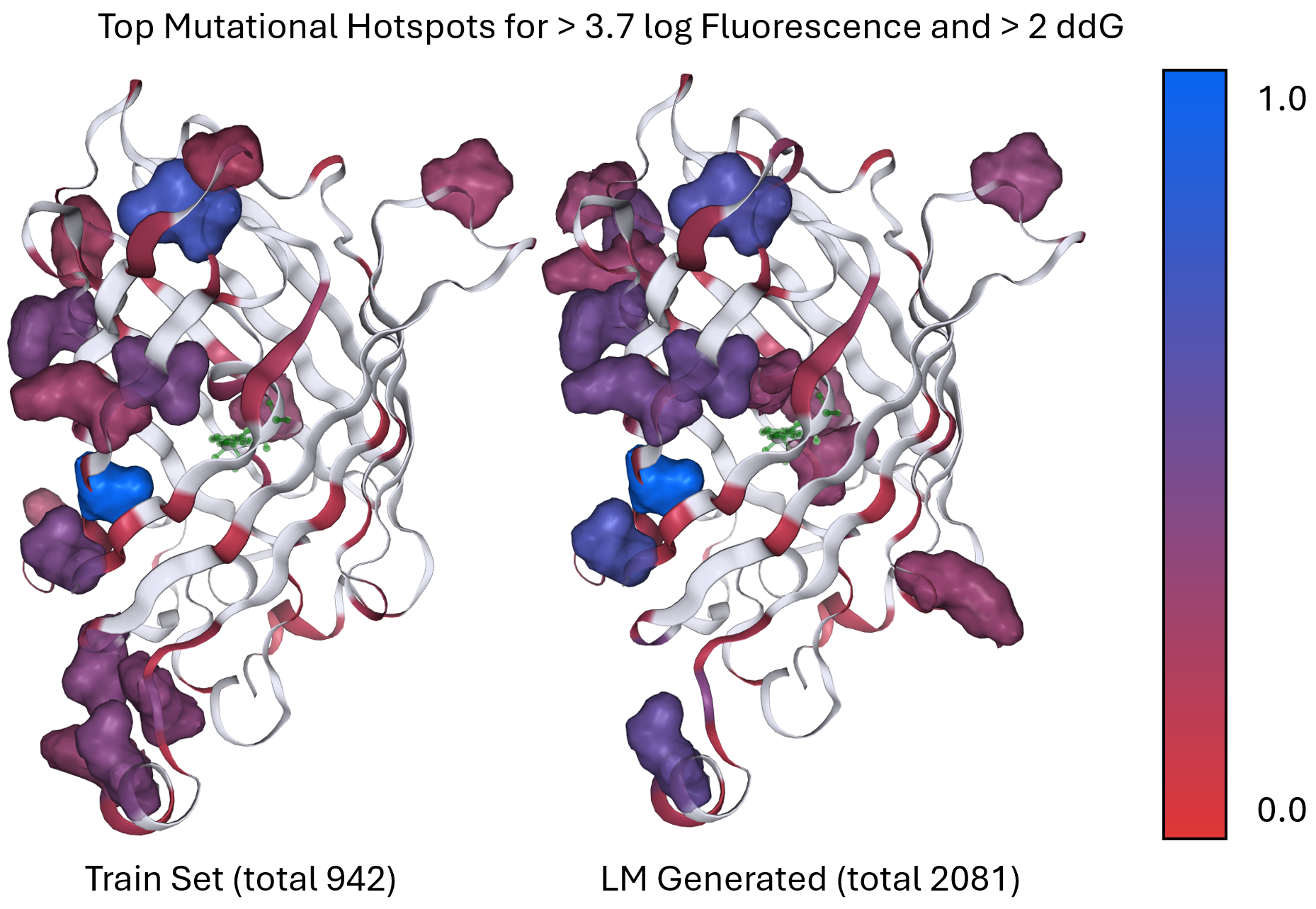}
  \caption{high fluorescence low stability}
\end{subfigure}
\caption{Comparing the top 15 mutational hotspots from the train set vs. LM predicted demonstrating that MACS can extrapolate beyond the mutational patterns seen during training.}
\label{fig:extra_protein_hotspots}
\end{figure}

\section{Limitations and Future Work}
{\framework} is an easy-to-implement framework to train domain-specific language models as fine-grained editors in an offline setting. However, there are a few limitations. Due to the offline nature of our method and our sampling strategy, it is unable to extrapolate well to regions within the multi-attribute space with low or no data points. To train good multi-attribute LM editors, {\framework} requires a good initial domain-specific pretrained language model. In our preliminary experiments with antibody generation task \citep{wong2023discovery}, a chemistry LM\footnote{\url{https://huggingface.co/ncfrey/ChemGPT-19M}} trained with {\framework} was not able to generate many novel candidates, likely due to its small size and poor data coverage. 

In the future, we aim to extend our method such that it can use both offline and on-policy samples to improve its performance and diversity in the fine-grained control task. Further research is also needed to support categorical and lexical constraints in {\framework}.

\end{document}